%% file: PaperForReview.tex
\documentclass[10pt,twocolumn,letterpaper]{article}

\usepackage[pagenumbers]{cvpr} % To force page numbers, e.g. for an arXiv version

\usepackage{graphicx}
\usepackage{amsmath}
\usepackage{amssymb}
\usepackage{pifont}
\usepackage{booktabs}
\usepackage[accsupp]{axessibility}  % Improves PDF readability for those with disabilities.

\usepackage[pagebackref,breaklinks,colorlinks]{hyperref}

\usepackage[capitalize]{cleveref}
\crefname{section}{Sec.}{Secs.}
\Crefname{section}{Section}{Sections}
\Crefname{table}{Table}{Tables}
\crefname{table}{Tab.}{Tabs.}

\def\cvprPaperID{4940} % *** Enter the CVPR Paper ID here
\def\confName{CVPR}
\def\confYear{2023}

\begin{document}

\title{3DQD: Generalized Deep 3D Shape Prior via Part-Discretized Diffusion Process}
\author{Yuhan Li\textsuperscript{1}\hspace{2mm}
Yishun Dou\textsuperscript{2}\hspace{2mm}
Xuanhong Chen\textsuperscript{1}\hspace{2mm}
Bingbing Ni\textsuperscript{1, 2$\dagger$}\hspace{2mm}
Yilin Sun\textsuperscript{1}\hspace{2mm}
Yutian Liu\textsuperscript{1}\hspace{2mm}
Fuzhen Wang\textsuperscript{1}\\
\textsuperscript{1}Shanghai Jiao Tong University, Shanghai 200240, China \qquad \textsuperscript{2}Huawei Hisilicon \\
{\tt\small \{melodious, nibingbing\}@sjtu.edu.cn}\\
{\small \url{https://github.com/colorful-liyu/3DQD}}
}

\maketitle

%%%%%%%%% ABSTRACT
\begin{abstract}
 \vspace{-3mm}
We develop a generalized 3D shape generation prior model, tailored for multiple 3D tasks including unconditional shape generation, point cloud completion, and cross-modality shape generation, \etc. On one hand, to precisely capture local fine detailed shape information, a vector quantized variational autoencoder (VQ-VAE) is utilized to index local geometry from a compactly learned codebook based on a broad set of task training data. On the other hand, a discrete diffusion generator is introduced to model the inherent structural dependencies among different tokens. In the meantime, a multi-frequency fusion module (MFM) is developed to suppress high-frequency shape feature fluctuations, guided by multi-frequency contextual information.
The above designs jointly equip our proposed 3D shape prior model with high-fidelity, diverse features as well as the capability of cross-modality alignment, and extensive experiments have demonstrated superior performances on various 3D shape generation tasks.
\end{abstract}

\newcommand{\customfootnotetext}[2]{{% Group to localize change to footnote
  \renewcommand{\thefootnote}{#1}% Update footnote counter representation
  \footnotetext[0]{#2}}}% Print footnote text
\customfootnotetext{${\dagger}$}{Corresponding author: Bingbing Ni.}

%%%%%%%%% BODY TEXT
 \vspace{-5mm}
\section{Introduction}
 \vspace{-2mm}
%%background
While pre-trained 2D prior models~\cite{karras2019style, rombach2021ldm} have shown great power in various downstream vision tasks such as image classification, editing and cross-modality generation, \etc., their counterpart 3D prior models  which are generally beneficial for three-dimensional shape generation tasks have NOT been well developed, unfortunately.
On the contrary, the graphics community has developed a number of task-specific pre-trained models, tailored for unary tasks such as 3D shape generation~\cite{yang2019pointflow, luo2021dpm, hui2022neural}, points cloud completion~\cite{zhou2021pvd, yu2021pointr, zhou2022seedformer} and conditional shape prediction~\cite{fu2022shapecrafter, liu2022towards, mittal2022autosdf, wang2018pixel2mesh}.  Since above individual 3D generative representations do NOT share common knowledge among tasks, to migrate a trained 3D shape network from one task to another related one requires troublesome end-to-end model re-work and training resources are also wasted.
For instance, a good shape encoding of ``chairs" based on a general prior 3D model could benefit shape completion of a given partial chair and text-guided novel chair generation.

\begin{figure}[t]
 \vspace{-5mm}
  \centering
   \includegraphics[width=\linewidth]{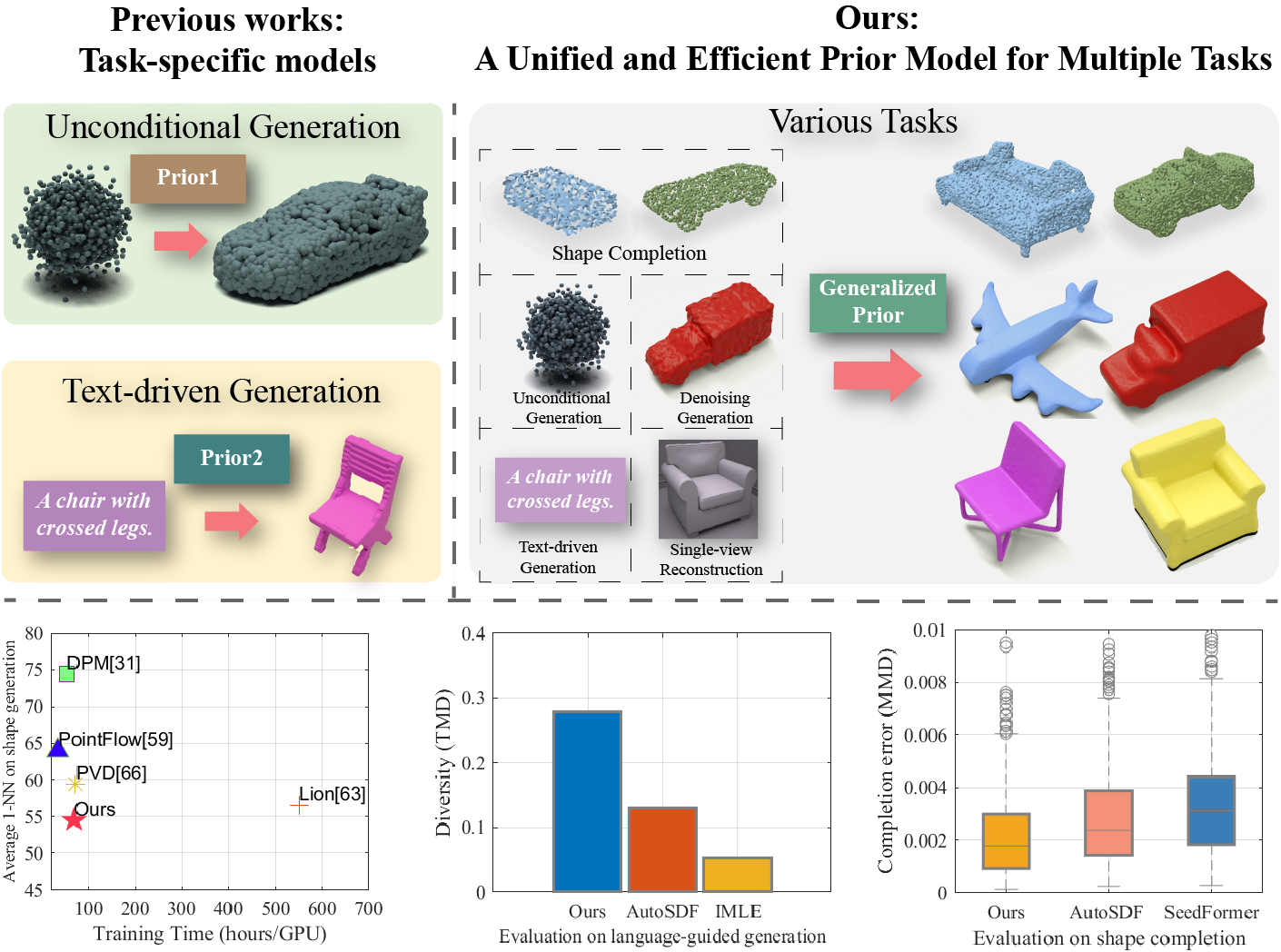}
   \caption{Our shape generation model is a unified and efficient prior model to produce high-fidelity, diverse results on multiple tasks, while most previous approaches are task-specific.}
   \label{motivation}
    \vspace{-5mm}
\end{figure}

%% chanlenge
This work aims at designing a unified 3D shape generative prior model, which serves as a generalized backbone for multiple downstream tasks, \ie, with few requirements for painstaking adaptation of the prior model itself.
Such a general 3D prior model should possess the following \emph{good} properties. On one hand, it should be \emph{expressive} enough to generate \textbf{high-fidelity} shape generation results with fine-grained local details. On the other hand, this general prior should cover a large probabilistic support region so that it could sample \textbf{diverse} shape prototypes in both conditional and unconditional generation tasks. 
Moreover, to well support \textbf{cross-modal} generation, \eg, text-to-shape, the sampled representation from the prior model should achieve good semantic consistency between different modalities, making it easy to encode and match partial shapes, images and text prompts.

%%problem
The above criteria, however, are rarely satisfied by existing 3D shape modeling approaches. Encoder-decoder based structures~\cite{yu2021pointr, zhou2022seedformer} usually focus on dedicated tasks and fail to capture diverse shape samples because the generation process mostly relies on features sampled from deterministic encoders. On the contrary, probabilistic models such as GAN~\cite{shu2019treegan, li2021sp}, Flow~\cite{yang2019pointflow} and diffusion models~\cite{luo2021dpm, hui2022neural, zhou2021pvd}, cannot either adapt to multiple tasks flexibly or generate high-quality shapes without artifacts~\cite{zeng2022lion}.
Note that recent models such as AutoSDF~\cite{mittal2022autosdf} and Lion~\cite{zeng2022lion} explore multiple conditional generative frameworks. However, AutoSDF~\cite{mittal2022autosdf} is defective in diversity and shows mode collapse in unconditional and text-guided generation, while training and inference of Lion~\cite{zeng2022lion} are costly.

\begin{figure*}[t]
 \vspace{-5mm}
  \centering
   \includegraphics[width=\linewidth]{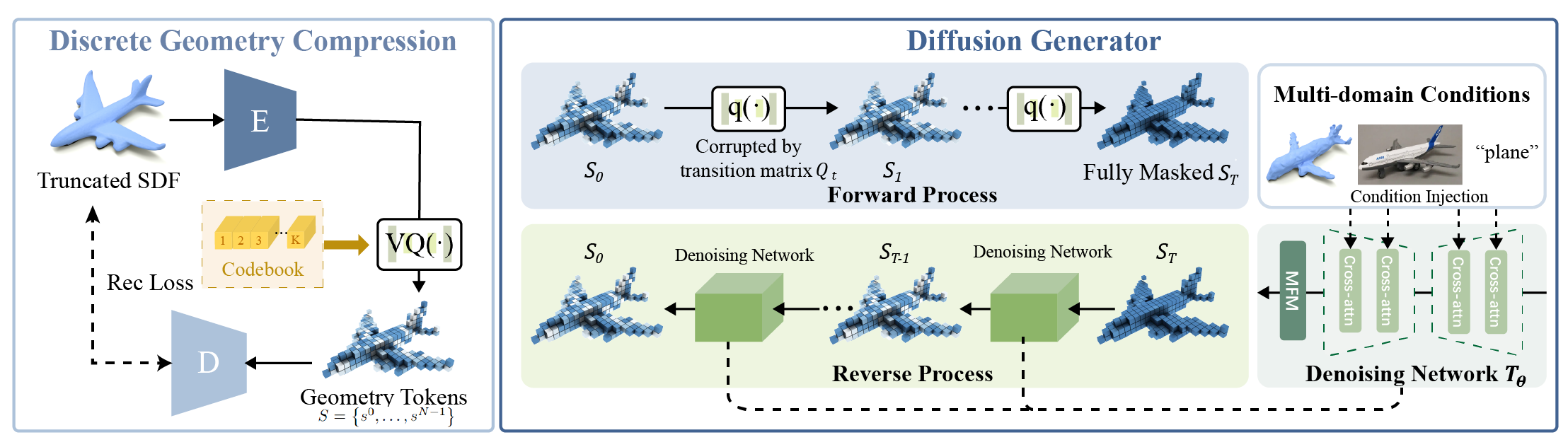}
   \caption{Overall framework of our method. VQ-VAE encodes shapes into geometry tokens with compact representation. Then diffusion generator models the joint distribution of tokens by reversing a forward diffusion process that corrupts the inputs via a Markov chain.}
   \label{framework}
    \vspace{-5mm}
\end{figure*}

%% My innovation
In view of above limitations, this work proposes an efficient \textbf{\emph{3D-Disassemble-Quantization-and-Diffusion (3DQD)}} prior model to simultaneously address above challenges in 3D shape generation: high-fidelity, diversity, and good cross-modality alignment ability, as shown in \cref{motivation}.
This 3DQD prior is a unified, probabilistic and powerful backbone for both unconditional shape generation and multiple conditional shape completion applications.
On one hand, instead of using holistic codes for whole-shape, we adopt a vector quantized variational autoencoder~\cite{van2017vqvae} (VQ-VAE) that learns a compact representation for \emph{disassembled} local parts of the shape, which is expected to well represent diverse types of local geometry information out of a broad set of training shapes from shared tasks, forming a generalized part-level codebook.
Note that this disassembled and quantized representation effectively eliminates the intrinsic structural bias between different modalities, and thus is able to perform good cross-modality data alignment on conditional generation tasks.
On the other hand, a discrete diffusion generator~\cite{austin2021structured, gu2022vector} with reverse Markov chain is introduced to model the inherent semantic dependencies among geometric tokens. Namely, the forward process corrupts the latent variables with progressively increasing noise, transferring parts of variables into random or masked ones; and the reverse process gradually recovers the variables towards the desired data distribution by learning the structural connections among geometry tokens.
It is worth mentioning that random corruption in the forward process facilitates diverse samples, while discrete embedding results in a dramatic cost-down in computational budget, with stable and iterative sampling. 
%It is worth noting that the above discrete diffusion generator gradually [estimates the probability density of geometry tokens step-by-step, where the random corruption leads to diverse results.
Furthermore, during shape generation we introduce \textbf{Multi-frequency Fusion Module} (MFM) to suppress high-frequency outliers, encouraging smoother and high-fidelity samples.

%%experiment
Comprehensive and various downstream experiments on shape generation tasks demonstrate that the proposed 3DQD prior is capable of helping synthesize high-fidelity, diverse shapes efficiently, outperforming multiple state-of-the-art methods. Furthermore, our 3DQD prior model can serve as a generalized backbone with highly competitive samples for extended applications and cross-domain tasks requiring NO or little tuning.

 \vspace{-2mm}
\section{Related Works}
 \vspace{-2mm}
\noindent
\textbf{Diffusion models.}
Recently, the denoising diffusion probabilistic model (DDPM), has attracted great attention in the community~\cite{ho2020denoising, nichol2021glide, rombach2021ldm}, which utilizes a Markov chain to convert the noise distribution to the data distribution. Models under this family have achieved strong results on image generation~\cite{ho2020denoising, nichol2021improved,  ho2022cascaded, dhariwal2021diffusion, ashual2022knn, saharia2022palette} and multiple conditional generation tasks~\cite{saharia2022image, rombach2021ldm, santos2022face, choi2021ilvr, meng2021sdedit, preechakul2021diffusion, chen2020wavegrad}. Meanwhile, other researchers investigated the discrete diffusion models. Argmax Flow~\cite{hoogeboom2021argmax} first introduces a diffusion model directly defined on discrete categorical variables, while D3PM~\cite{austin2021structured} generalizes it by going beyond corruption processes with uniform transition probabilities. VQ-Diffusion~\cite{gu2022vector, tang2022improved} encodes images into discrete feature maps to train the diffusion, presenting comparable results in text-to-image synthesis with the state-of-the-art models~\cite{nichol2021glide, ramesh2022hierarchical}.

\noindent
\textbf{3D Shape generation.} 
Traditional shape generation methods~\cite{yang2018foldingnet, groueix2018papier, li2021sp} mitigate generation process by mapping a primitive matrix to a point cloud with transformation. Some recent works~\cite{yang2019pointflow, cai2020learning, luo2021dpm, zhou2021pvd} consider point clouds as samples from a distribution by different probabilistic models. Almost all of them cannot cope with conditional input and multi-modality data. Some conditional works~\cite{chen2019unpaired, tchapmi2019topnet, yuan2018pcn, zhang2021unsupervised, yu2021pointr, zhou2022seedformer} focus on completing full shapes from partial inputs on point clouds with task-specific architectures to encode inputs and decode the global features. Others explore single-view reconstruction~\cite{mandikal20183d, wu2020pq, wang2018pixel2mesh, wang20193dn, wu2018learning, shi2021geometric} and text-driven generation~\cite{chen2018text2shape, mittal2022autosdf, liu2022towards, fu2022shapecrafter} to learn a joint condition-shape distribution with deterministic process. However, most conditional generation methods fail in capturing the multiple output modes~\cite{mittal2022autosdf} and adapting to various tasks.

Most related to our work, PVD~\cite{zhou2021pvd} and Lion~\cite{zeng2022lion} both use a diffusion model for diverse generation. Their models work on latent embedding space, leading to either difficulty in multi-modality applications (\ie, denosing)~\cite{zeng2022lion} or time-consuming training and inference. Inspired by discrete approaches~\cite{esser2021taming}, we introduce VQ-VAE~\cite{van2017vqvae} to learn a compact representation, followed by a discrete diffusion generator, to reduce computational overhead and align cross-domain data. Our model is a superior and efficient prior model for shape generation compared with PVD~\cite{zhou2021pvd} and Lion~\cite{zeng2022lion}, and shows more diversity than AutoSDF~\cite{mittal2022autosdf}.

 \vspace{-3mm}
\section{Methodology}
 \vspace{-2mm}
 
In this section, we introduce in detail of our proposed general 3D shape prior model (3DQD) for various 3D generation tasks. The architecture is visualized in \cref{framework}.
To begin with, we present our shape encoding scheme based on part-level discretization using VQ-VAE~\cite{van2017vqvae} style methods, for its advantages in representation compactness and consistency among different tasks.
Then, a novel diffusion process based prior is developed according to this discrete part based shape representation with expressive and diversified object structural modeling, forming a good basis for 3D generation downstream applications.
Moreover, a novel multi-frequency fusion module is introduced for enhancing fine-grained 3D surface modeling.

\subsection{Discrete Part-based Geometry Encoding}
 \vspace{-2mm}

A proper representation (\ie, shape encoding) is essential for effective shape distribution model design. Previous coding schemes~\cite{zhou2021pvd, luo2021dpm} based on continuous embedding representation learning suffer from drawbacks of large computational consumption and heavy task dependence, rendering difficulty in 3D conditional shape generation, especially for multi-task settings. While most encoder-based methods (\ie Lion~\cite{zeng2022lion}) choose to embed shapes into continuous space via VAE~\cite{kingma2013vae}, in this work, we propose to circumvent this drawback by introducing Patch-level VQ-VAE (P-VQ-VAE)~\cite{van2017vqvae, mittal2022autosdf} on Truncated-Signed Distance Field~\cite{xu2019disn, jiang2020sdfdiff} (T-SDF) to learn a compact and efficient representation. The motivation is obvious. All 3D objects are composed of geometrically similar local components, and thus a good local atomic representation can be shared among different objects and diverse tasks.% As a consequence
Besides, using discrete codebook to index each local part results in a dramatic cost-down in searching discrete feature space. As it's flexible to encode inter-dependencies between parts, this encoding scheme provides a compact way for expressing highly diverse intrinsic structural variations. Moreover, local atomic indexing potentially offers unified and well-aligned local representation among different modalities, \eg images and text prompts, which greatly facilitate cross-domain conditional shape generation.

Formally, the proposed P-VQ-VAE consists of an encoder $E$, a decoder $D$ and codebook entries $\mathcal{Z}  \in \mathbb{R }^{K\times n_z} $, which contains a finite number of embedding vectors, where $K$ is the size of codebook and $n_z$ is the dimension of vectors. Following AutoSDF~\cite{mittal2022autosdf}, given a T-SDF input $X \in \mathbb{R }^{H\times W\times D}$, the whole 3D shape is divided into partial regions $X'=\left \{ x'^{0}, \dots, x'^{N-1} \right \}$, where $x'^i\in \mathbb{R }^{h\times w\times d}$ and $N$ is the number of regions.  Afterwards, each part is vectorized by the encoder as $z^{i} = E(x'^{i}) \in \mathbb{R }^{n_z}$. Superscript $i$ is its spatial location.
Finally, we obtain a spatial collection of quantized shape tokens $Z_q=\left \{ z_{q}^0, \dots, z_{q}^{N-1} \right \}$ by a further quantization step as $z_{q}^i = VQ(z^{i}) \in \mathbb{R }^{n_z}$, and their corresponding codebook indexes $S=\left \{ s^{0}, \dots, s^{N-1} \right \}$, where $s^i\in \left \{ 0, \dots, K-1 \right \}$. $VQ$ is a spatial-wise quantizer which maps partial shape $z^i$ into its closest codebook entry in $\mathcal{Z} $.

\subsection{3D Shape Prior: Discrete Diffusion Model}
 \vspace{-2mm}
 
Given the above P-VQ-VAE encoding (\emph{i.e.}, discrete part geometry tokens $Z_q\in \mathbb{R }^{N\times n_z}$ and corresponding indexes $S=\left \{ s^0, \dots, s^{N-1} \right \}$), to build the 3D shape prior is equivalent to modeling the \textbf{joint probability distribution} of all local shape codes in the latent space. In essence, a sampling from this prior distribution reveals a certain intrinsic structure of a 3D shape, by considering the inter-relationship geometric organization among local shape codes, \emph{i.e.}, how to spatially combine local parts into various 3D objects.
In this sense, a good prior model, should not only provide a probabilistic support region as wide as possible (\ie possessing sufficient shape diversity), but also be general enough to deal with different downstream 3D shape tasks. For example, a chair with 4 legs can be transformed into a chair with wheels easily by the knowledge learned in generation, without training on target editing tasks.

The joint distribution of partial components and conditions are usually learned with autoregressive models in previous works, \eg, DALL-E~\cite{ramesh2021dalle}, Taming~\cite{esser2021taming}, ShapeFormer~\cite{yan2022shapeformer} and AutoSDF~\cite{mittal2022autosdf}, along with sequential non-Markov-chain styled generation process: $ {\textstyle \prod_{i=1}^{N}p_{\theta }(s^i|s^1, \dots ,s^{i-1})} $. However, this scheme has several drawbacks. 1) \textbf{Error Accumulation}: Geometry tokens are predicted one by one, therefore errors induced in the earlier sampling timesteps will never be corrected and contaminate the subsequent generation steps; 2) \textbf{Un-optimal Generation Order}: Most autoregressive models perform an unidirectional prediction process, \eg left-to-right, bottom-to-top, front-to-back or random orders, which obviously ignores the underlying complex 3D structure; and 3) \textbf{Lack of Diversity}: Deterministic transformers are usually instantiated as backbones of autoregressive models~\cite{yan2022shapeformer, mittal2022autosdf}, and without sufficient randomness injection they easily lean towards mode collapse (\eg, highly similar completion results given a partial shape), especially in condition-driven tasks.

In view of these limitations, we develop a discrete diffusion generator~\cite{tang2022improved, gu2022vector} to iteratively sample in the time domain with all local part codes updated simultaneously at each timestep (\ie , instead of sampling each spatial location one by one). In this way, diffusion generator is able to get rid of fixed unidirectional generation and update all partial geometry with long-range dependencies simultaneously, which enhances structural expressivity of the learned shape distribution.
Within this framework, earlier samples are rechecked multiple times during iterations, reducing the likelihood of being confused. In addition, random corruption on shapes in diffusing process also leads to a great diversity of generated results.
The forward sampling process for shape generation and the backward sampling process for training the proposed discrete diffusion model are introduced in detail as follows.

%-----------------------------------------------------------------------------------
 \vspace{-2mm}
\subsubsection{Forward Diffusing Process and Corruption.}
 \vspace{-2mm}
 
The forward diffusion process gradually adds noise to tokenized 3D geometric input $S_0=\left \{ s_0^{0}, \dots, s_0^{N-1} \right \}$ (superscripts are locations and subscripts are timesteps) via a Markov chain $q(s^i_{1:T}|s^i_0)={\textstyle \prod_{t=1}^{T}q(s^i_t|s^i_{t-1})} $, where each token is randomly replaced into noisy index $s^i_{1:T}=s^i_1,s^i_2,\dots ,s^i_T$. Without introducing confusion, we omit superscripts $i$ in the following description.  After $T$ timesteps, each token in the entire map will be completely corrupted into $s_T$, \ie, a non-sense index. The learned reverse process $p_{\theta }(s_{0:T})=p(s_T) {\textstyle \prod_{t=1}^{T}p_{\theta}(s_{t-1}|s_t)} $ gradually removes the added noise on the random variables $s_t$, namely to generate a 3D shape from random noise.

While continuous diffusion process models~\cite{ho2020denoising, zhou2021pvd, zeng2022lion} employ Gaussian noise in the forward process; in discrete diffusion process, we use transition matrices $\left [ Q_t \right ] _{i,j}=q(s_t=j|s_{t-1}=i)$ characterized by uniform transition probabilities to describe the corruption from $s_{t-1}$ to $s_{t}$. As a result, all local part codes can be transformed to any other shape codes with the same transition probability, producing complex 3D structure representations. With transition matrix $\left [ Q_t \right ]$ and one-hot encoding of $s_t$, we define the forward Markov chain as:
\begin{small}
\vspace{-2mm}
\begin{equation}
q(s_t|s_{t-1}) = \mathbf{\Psi} (s_t;p=s_{t-1}Q_t),
\label{q(st|st-1)}
\end{equation}
\end{small}
where $\mathbf{\Psi} (s_t;p)$ is a categorical distribution over the one-hot row vector $s_t$ sampled with probability $p$, and $s_{t-1}Q_t$ is computed by a row vector-matrix product. 
Accordingly, we derive the posterior by Markov chain iteratively from $s_0$ as:

\begin{small}
\vspace{-5mm}
\begin{equation}
q(s_t|s_0) = \mathbf{\Psi} (s_t;p=s_{0}\overline{Q_t})\text{,~~with }\overline{Q_t}=Q_1\cdots Q_t,
  \label{q(st|s0)}
\end{equation}
\begin{equation}
\begin{split}
q(s_{t-1}|s_t, s_0) &= \frac{q(s_t|s_{t-1}, s_0)q(s_{t-1}|s_0)}{q(s_t|s_0)} \\
&= \mathbf{\Psi} \left (  s_{t-1};p=\frac{s_tQ_t^\top \odot s_0\overline{Q}_{t-1} }{s_0\overline{Q}_t s_t^\top }\right ).
\end{split}
  \label{posterior}
\end{equation}
\end{small}

The transition matrix $Q_t$ reflects how to spatially re-organize 3D local primitives and it determines the degree of freedom in deformation. In order to assist the network to locate what needs to be fixed quickly, we introduce an additional special token, [\emph{MASK}] token~\cite{gu2022vector}. So the codebook entries are composed of $K+1$ states: $K$ geometry tokens and one [\emph{MASK}] token. $Q_t\in \mathbb{R}^{(K+1) \times (K+1)}$ can be formulated as:

\begin{small}
\vspace{-5mm}
\begin{equation}
  \left [ Q_t \right ] = \
  \begin{bmatrix}\alpha _t-\frac{K-1}{K}\beta _t & \frac{\beta _t}{K} & \frac{\beta _t}{K} & \cdots & \gamma_t
 \\\frac{\beta _t}{K} & \alpha _t-\frac{K-1}{K}\beta _t & \frac{\beta _t}{K} & \cdots & \gamma_t
 \\\frac{\beta _t}{K} & \frac{\beta _t}{K} & \alpha _t-\frac{K-1}{K}\beta _t & \cdots & \gamma_t
 \\\vdots  & \vdots  & \vdots  & \ddots  & \vdots 
 \\0 & 0 & 0 & \cdots & 1
\end{bmatrix},
\label{qt_ij}
\end{equation}
\end{small}
with $\alpha _t = 1 - \gamma _t$, which means each token has a probability of $1-\gamma _t-\frac{K-1}{K}\beta _t$ to remain unchanged, while with a probability of $\frac{\beta _t}{K}$ being transited equally into other $K-1$ geometry categories and $\gamma _t$ into [\emph{MASK}] state.

%-----------------------------------------------------------------------------------

\begin{figure}[t]
  \centering
   \includegraphics[width=\linewidth]{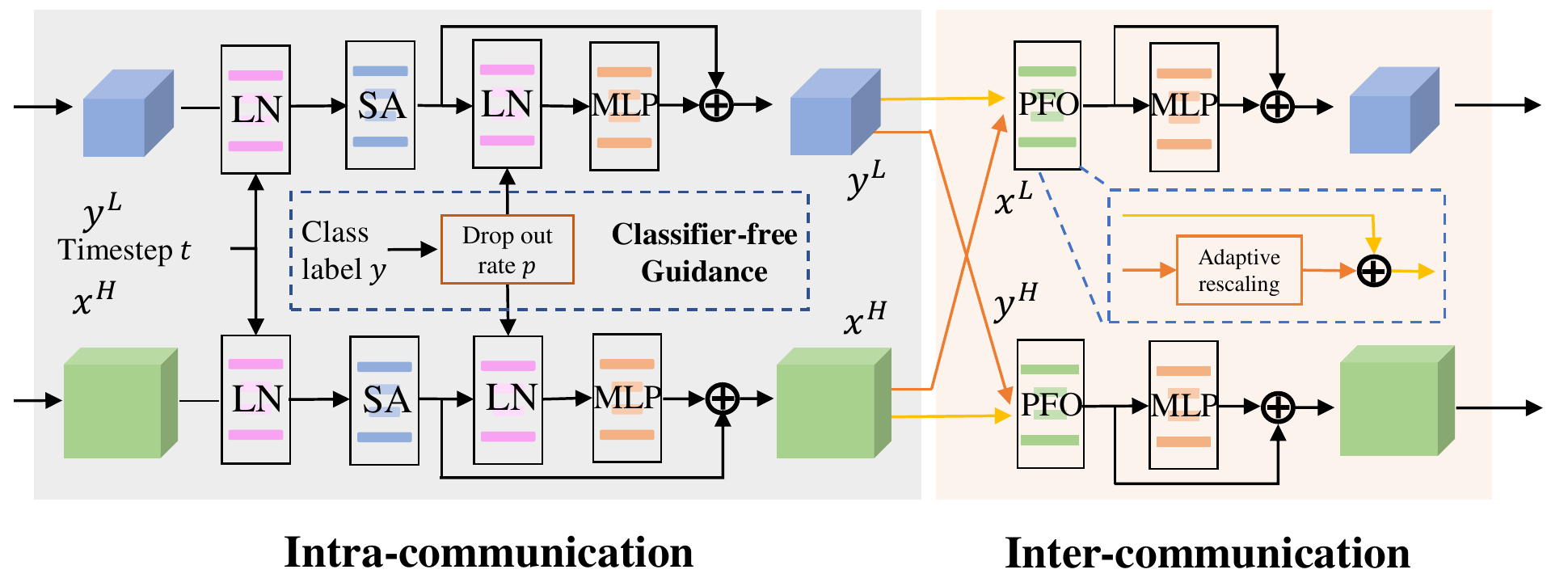}
   \caption{Information from detailed shapes and condensed features are intra- and inter-communicated in MFM to remove high-frequency components from uniform noise, where timestep $t$ and classifier-free guidance are modulated with AdaLayerNorm.}
   \label{MFM}
    \vspace{-5mm}
\end{figure}

 \vspace{-2mm}
\subsubsection{Reverse Sampling Process and Model Learning}
 \vspace{-2mm}
 
To recover original 3D grids of tokens from $s_t$, we follow Ho \etal~\cite{ho2020denoising} and Hoogeboom \etal~\cite{hoogeboom2021argmax}, and train a denoising transformer $T_{\theta }$ to directly estimate the noise-removed shape distribution $p_{\theta }(\hat{s}_0 |s_t)$. Then the posterior transition distribution can be obtained with $q(s_{t-1}|s_t, s_0)$ as:

\begin{small}
\vspace{-2mm}
\begin{equation}
p_{\theta }(s_{t-1}|s_t) = \sum_{\hat{s}_0} q(s_{t-1}|s_t, \hat{s}_0)p_{\theta}(\hat{s}_0|s_t).
  \label{p(st-1|st)}
\end{equation}
\end{small}

The evidence lower bound (ELBO) associated with this model then decomposes over the discrete timesteps as:
\begin{small}
 \vspace{-1mm}
\begin{equation}
\begin{split}
-\log &p_{\theta }(\hat{s_0}) \le \mathbb{KL}(q(s_T|s_0)|p(s_T))\\
&+\sum_{t=1}^{T}\mathbb{E}_{q(s_t|s_0)}\mathbb{KL}(q(s_{t-1}|s_t, s_0)|p_{\theta}(s_{t-1}|s_t)).
\end{split}
  \label{ELBO}
  \vspace{-3mm}
\end{equation}
\end{small}

For the first term of ELBO has no relation with the learnable network ($p(s_T)$ is an initial noise distribution), the common way to minimize ELBO is to optimize the posterior $p_{\theta}(s_{t-1}|s_t)$ in terms of~\cref{p(st-1|st)}. Due to the direct prediction of neural network is original shapes $p_{\theta}(\hat{s}_0|s_t)$, we set an auxiliary training objective with the distance of recovered original shapes apart from the posterior, yielding the following loss function:
\begin{small}
 \vspace{-3mm}
\begin{equation}
L=-\mathbb{E} _{q(s_0)q(s_t|s_0)}\left [\underset{L_{main}}{\underbrace{\log p_{\theta}(s_{t-1}|s_t)}} +\underset{L_{aux}}{\lambda \underbrace{\log p_{\theta}(\hat{s}_0|s_t)}} \right ],
  \label{Loss}
  \vspace{-2mm}
\end{equation}
\end{small}
where $\lambda $ balances the weight of the auxiliary loss.

\noindent
\textbf{Classifier-free Guidance.} To trade off between precision and diversity, Ho \& Nichol \etal~\cite{ho2022classifier, nichol2021glide} propose classifier-free guidance, which does not require a separately trained classifier model. During training, the class label $y$ in diffusion model has a fixed probability (drop out rate $p$ ) to be replaced with the empty label $\emptyset$. During inference, the model prediction is thus adjusted in the direction of $p_{\theta}(\hat{s}_0|s_t, y)$ away from $p_{\theta}(\hat{s}_0|s_t, \emptyset)$ as:

\begin{small}
 \vspace{-4mm}
\begin{equation}
p_{\theta}(\hat{s}_0|s_t) = (1+w)\cdot p_{\theta}(\hat{s}_0|s_t, y)-w\cdot p_{\theta}(\hat{s}_0|s_t, \emptyset),
  \label{cf-guide}
\end{equation}
\end{small}
with $w\ge 0$ controlling the guidance scale.

In this work, we also employ classifier-free guidance on multi-category generation tasks ( \eg, shape completion), which is proved to enrich diversity of samples without apparently affecting the fidelity. We set drop out rate $p=0.5$ and guidance weight $w=0.5$.

 \vspace{-2mm}
\subsubsection{Multi-frequency Fusion Module}
\label{mainbodyMFM}
 \vspace{-2mm}
 
In the meantime, we observe noisy surfaces with outliers from generated shapes mainly introduced by \emph{categorical corruption} in discrete diffusion models. Namely, our \emph{categorical corruption} has equal probability to transit a token into any completely irrelevant shape tokens~\cite{austin2021structured}, and it is hard to recover the correct category of this shape token without looking into its contextual information, \eg, its adjacent tokens.
Note that in continuous diffusion models, the added Gaussian noise only brings up \emph{soft} transition which still preserves part of the original shape information. Therefore, our transition matrix $Q_t$ with uniform noise imports more high-frequency noise and outliers than Gaussian diffusion~\cite{ho2020denoising} and autoregressive model~\cite{van2017vqvae, esser2021taming}.

To remedy this issue, inspired by the observation that corrupted tokens' neighbors are still possible to remain unchanged and provide dependencies, we develop a Multi-frequency Fusion Module (MFM) which looks into local contextual regions and extracts low-frequency components by downsampling to suppress high-frequency outliers, encouraging smoother and high-fidelity samples.
As shown in~\cref{MFM}, we split the detailed shape token embedding $x^H$ explicitly with the condensed features $y^L$ from its neighbors, which is obtained by down-sampling within its local receptive field. Each component is further sent into intra- and inter-relationship part: $[x^{H\longrightarrow H\longrightarrow L} ]$ and $[y^{L\longrightarrow L\longrightarrow H} ]$, respectively, and communication of information is realized. Specifically, we employ self-attention for intra-frequency update, and Pair-wise Fusion Operator (PFO) for communication between two components. We examine two ways of communication ( \ie cross-attention~\cite{chen2021crossvit}, residual add~\cite{he2016deep}). Note that residual add 
\begin{small}
\vspace{-1mm}
\begin{equation}
x^H+f(x^H+\mathcal{P}(y^L)),
  \label{res_add}
\end{equation}
\vspace{-1mm}
\end{small}
is set as default Pair-wise Fusion Operator in experiment, where $f$ is fully connected layers, and $\mathcal{P}$ means pair-wise alignment. We employ 3 MFM layers cascaded at the end of the denoising transformer, as filters to suppress high-frequency outliers, as validated in experiment.

\begin{figure*}[t]
 \vspace{-3mm}
  \centering
   \includegraphics[width=\linewidth]{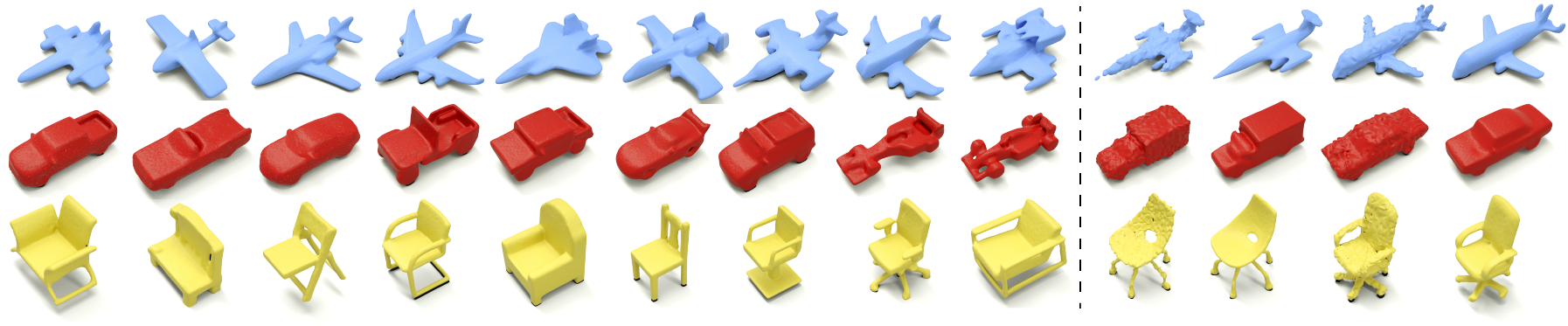}
   \caption{Our approach synthesizes high-quality and diverse shapes with smooth surfaces and complex structures on multiple tasks. \textit{Left}: Unconditional shape generation. \textit{Right}: Shape denoising without fine-tuning.}
   \label{uncond_denoise}
    \vspace{-5mm}
\end{figure*}

\begin{figure*}[t]
\centering
\begin{minipage}[t]{0.65\linewidth}
\centering
\includegraphics[width=\linewidth]{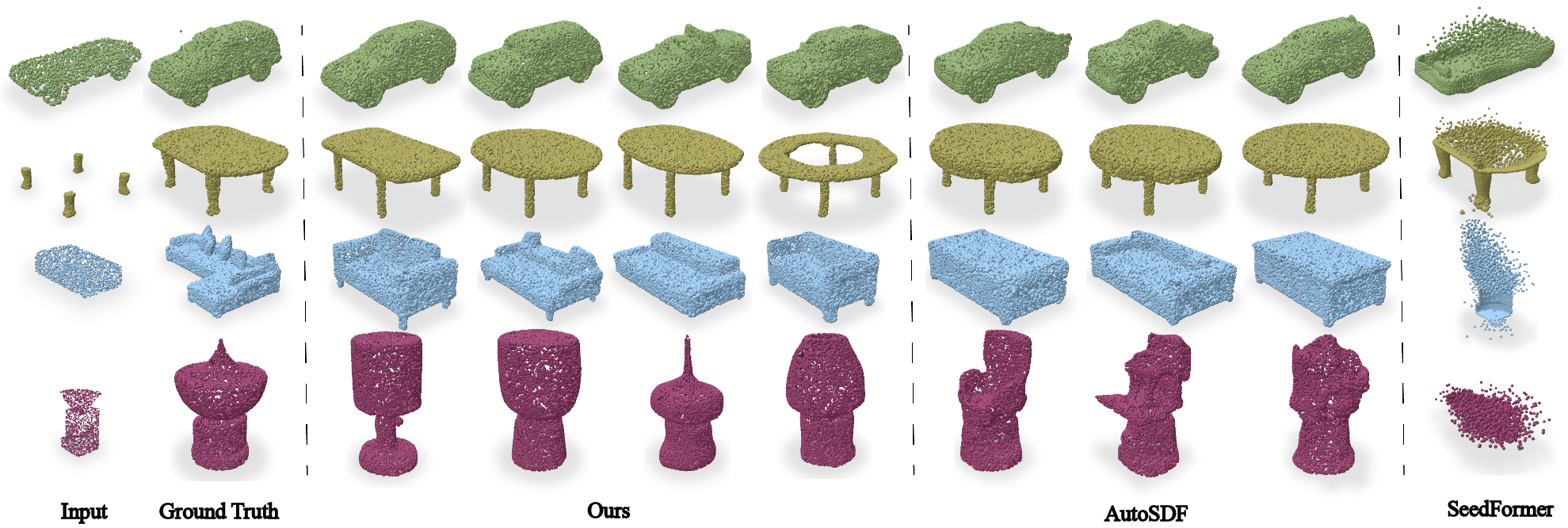}
\caption{Shape completion comparison. Our method yields most high-fidelity and diverse generation by the powerful and iterative modeling of joint distribution from diffusion, as well as generalized prior, even though only a small part is given in the last two rows.}
\label{completion_comp}
\end{minipage}
\begin{minipage}[t]{0.33\linewidth}
\centering
\includegraphics[width=\linewidth]{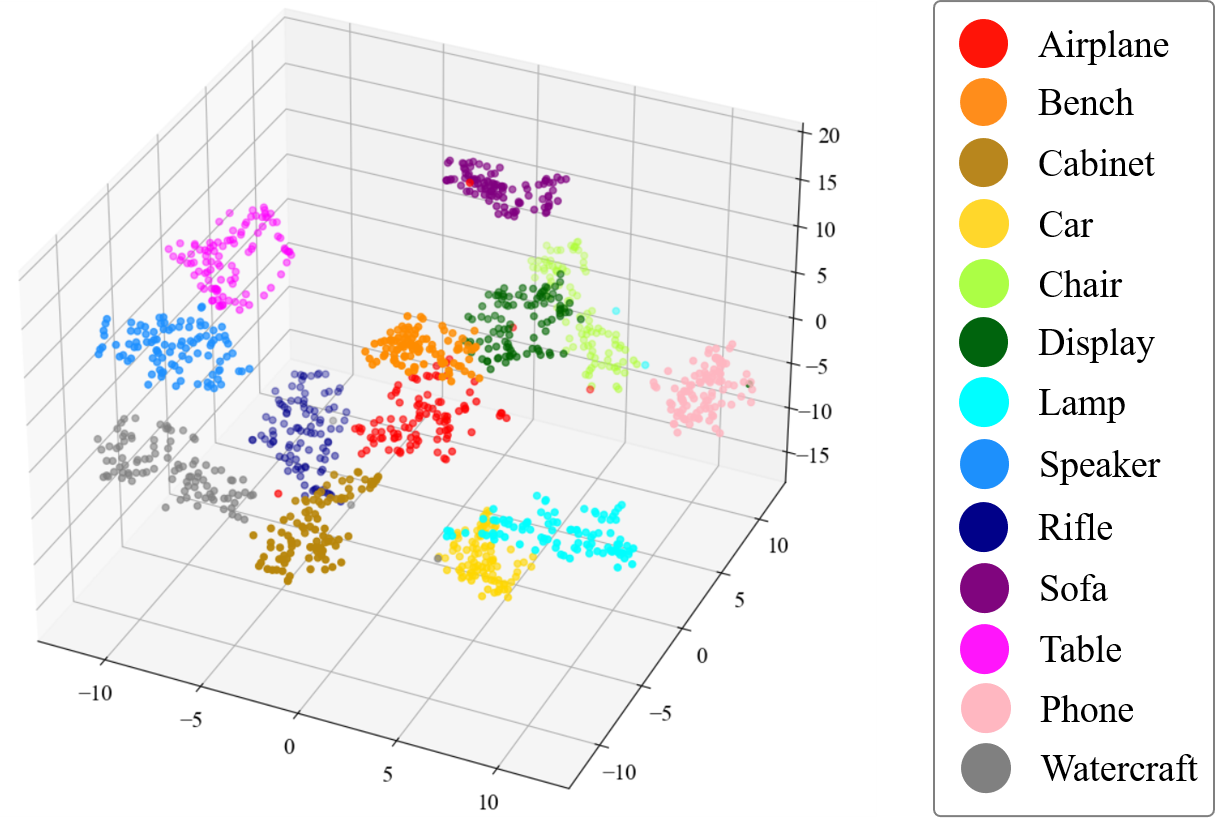}
\caption{We visualize the prior with T-SNE projections. Observe intra-class coherence and inter-class divergence.}
\label{tsne}
\end{minipage}
 \vspace{-5mm}
\end{figure*}

\section{Experiments}
\label{sec:Experiments}
 \vspace{-2mm}
In this section, we quantitatively and qualitatively evaluate our proposed unified 3DQD prior on three mainstream 3D shape generation tasks. We also discuss how 3DQD is extended for different relevant applications.

\subsection{3D Shape Generation}
\label{sec:uncond}
 \vspace{-2mm}
\noindent
\textbf{Data and evaluation.} We select three most popular categories from ShapeNet~\cite{chang2015shapenet}: \textit{Airplane}, \textit{Chair}, \textit{Car} as our main datasets for training. For input, we prepare the same T-SDF files as in DISN~\cite{xu2019disn} and follow the train-test split. Following previous works~\cite{zhou2021pvd, zeng2022lion}, we use 1-NNA as our main metric with both Chamfer distance (CD) and earth mover distance (EMD), measuring both shape generation quality and diversity. Limitations of other metrics are also discussed in our supplementary material.

\noindent
\textbf{Baselines and results.} We present the visual samples in~\cref{uncond_denoise} and quantitative  results in~\cref{tab:uncondition}. In comparison with baseline methods, we follow the same data processing and evaluation procedure as in PVD~\cite{zhou2021pvd}. Since our data is in the format of volumetric T-SDFs, for a fair comparison, we first transform T-SDFs into meshes, and then sample 2048 points for each mesh. It is noted that 3DQD outperforms all baselines with the best shape generation quality, due to the stable iterative generation and powerful joint distribution modeling capabilities of discrete diffusion generator, while MFM layers successfully suppress high-frequency outliers and improve the smoothness of results.

\begin{table}
  \centering
  \resizebox{\linewidth}{!}{
  \begin{tabular}{lcccccc}
  \toprule
               & \multicolumn{2}{c}{Airplane}    & \multicolumn{2}{c}{Chair}       & \multicolumn{2}{c}{Car}  \\  \cline{2-7} 
Method       & CD$\downarrow $             & EMD$\downarrow $            & CD$\downarrow $             & EMD$\downarrow $            & CD$\downarrow $             & EMD$\downarrow $     \\
\midrule
r-GAN~\cite{achlioptas2018learning}         & 99.84          & 96.79          & 83.69          & 99.70          & 94.46          & 99.01    \\
PointFlow~\cite{yang2019pointflow}     & 75.68          & 70.74          & 62.84          & 60.57          & 58.10          & 56.25    \\
SoftFlow~\cite{kim2020softflow}      & 76.05          & 65.80          & 59.21          & 60.05          & 64.77          & 60.09    \\
SetVAE~\cite{kim2021setvae}        & 76.54          & 67.65          & 55.84          & 60.57          & 59.94          & 59.94    \\
DPF-Net~\cite{klokov2020discrete}       & 75.18          & 65.55          & 62.00          & 58.53          & 62.35          & 54.48    \\
DPM~\cite{luo2021dpm}           & 76.42          & 86.91          & 60.05          & 74.77          & 68.89          & 79.97    \\
PVD~\cite{zhou2021pvd}          & 73.82          & 64.81          & 56.26          & 53.32          & \textbf{54.55} & 53.83   \\
\midrule
Ours  & \textbf{56.29} & \textbf{54.78} & \textbf{55.61} & \textbf{52.94} & 55.75         & \textbf{52.80}\\

\bottomrule

  \end{tabular}}
  \caption{Generation results on \textit{Airplane}, \textit{Chair}, \textit{Car} categories from ShapeNet~\cite{chang2015shapenet} using 1-NN$\downarrow $ as the metric.}
  \label{tab:uncondition}
   \vspace{-5mm}
\end{table}

\subsection{Shape Completion}
\label{sec:completion}
 \vspace{-2mm}
 
\noindent
\textbf{Data and evaluation.} We evaluate our method on the ShapeNet dataset with 13 categories using the train/test splits provided by DISN~\cite{xu2019disn}. As partial input is in the form of T-SDFs, we use the benchmark from AutoSDF~\cite{mittal2022autosdf}, which contains two different settings of observed shapes: 1) \textit{Bottom half} of ground truth as partial shape, and 2) \textit{Octant} with front, left and bottom half of ground truth as partial shape. For evaluation, we compute \textit{Total Mutual Difference} (TMD) of $N=10$ generated shapes for each input. We also report \textit{Minimum Matching Distance} (MMD) and \textit{Average Matching Distance} (AMD) which measure the minimum and average Chamfer distance from ground truth to the $N$ generated shapes.

\noindent
\textbf{Condition injection.} 
It's vital for us to explore how to infer missing shapes in terms of the partial observations (out of distribution) with a model trained with noisy complete shape tokens. To this end, we encode the inputs into codebook tokens and diffuse them at $k$ timestep into $\widetilde{s}_k$, to replace the full masked $s_T$ for iterative generation. So that the reverse process starts at $k$ timestep rather than $T$ timestep, with partial information kept in incompletely corrupted token maps. We set $k=\frac{1}{2}T$, as smaller $k$ value damages diversity in shape generation and larger $k$ reduces the fidelity of generation shapes.

 \begin{figure}[t]
  \centering
   \includegraphics[width=\linewidth]{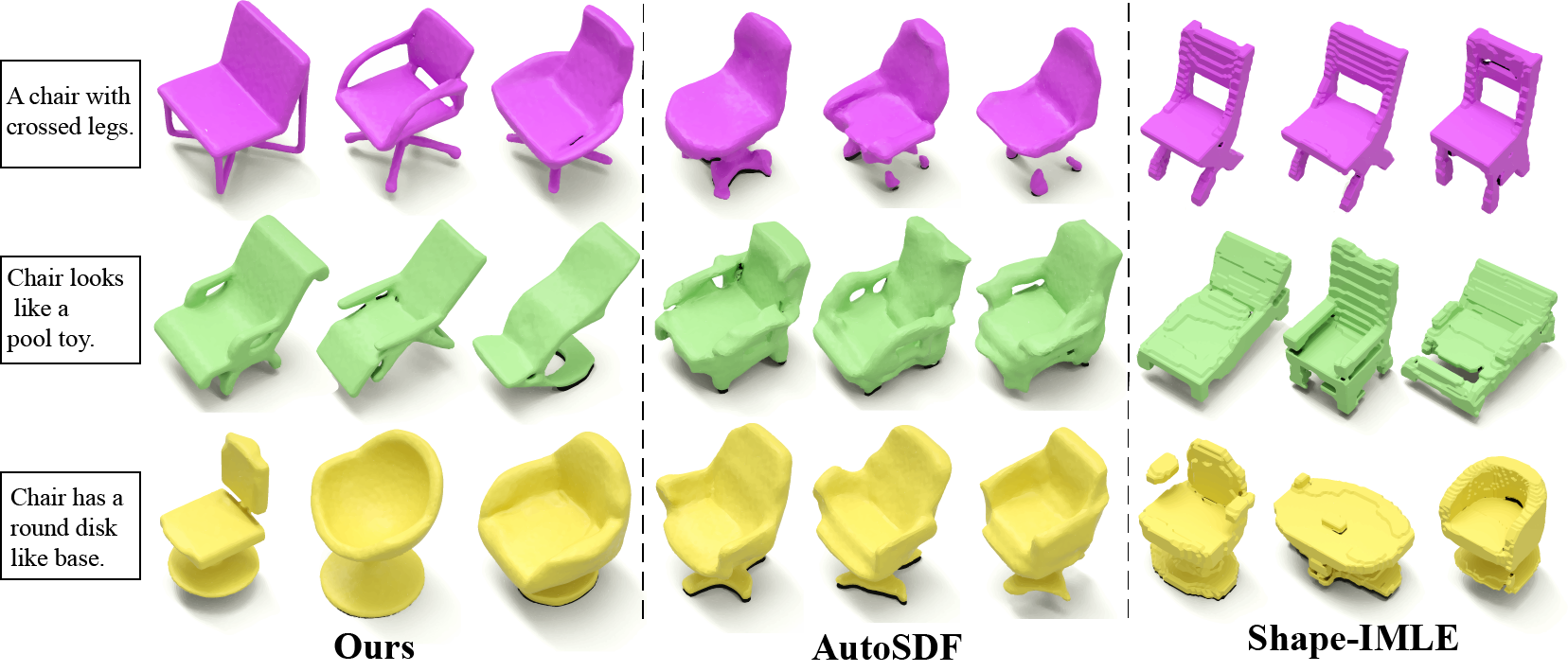}
   \caption{Comparative results for language-guided generation. Our model captures the reasonable variations from text prompts with smooth surfaces, while others produce unsatisfying results.}
   \label{text_3d}
    \vspace{-5mm}
\end{figure}

\noindent
\textbf{Baselines and results.} 
We compare 3DQD with three state-of-the-art shapes completion methods, including PoinTr~\cite{yu2021pointr}, SeedFormer~\cite{zhou2022seedformer} and AutoSDF~\cite{mittal2022autosdf}. The former two methods perform accurate supervised completion, while the latter is able to sample diverse shapes from a partial observation. A set of random held-out shapes from ShapeNet dataset with 13 categories are used to compare the performance. The number of objects in the set is $M=500$. Quantitative results are presented in~\cref{tab:completion} and several visualized results are shown in~\cref{completion_comp}. 
It is observed that our method surpasses all models in both diversity and fidelity of generation, thanks to the under-determined, multi-modality nature of diffusing process. 
It is also worth mentioning that although our model is trained only on completely noisy shape tokens, it successfully performs inference conditioned on local noisy tokens with rest regions fully masked (which it has never seen). In contrast, models which are specifically trained only for 3D completion fail this task, as illustrated in the right column of~\cref{completion_comp}. It also indicates that our model has learned prior knowledge about the structural relations between local shapes, and this joint prior can be extended to more applications as discussed in~\cref{sec:Extensions}.

\noindent
\textbf{Prior Analysis.} We further visualize our learned prior distribution to provide an in-depth analysis about how the underlying shape structure distributions for various object types are well modeled by 3DQD so that they could easily facilitate various tasks. For this purpose, we sample $N=100$ shape completion results from only an octant input for each object category, and then perform feature encoding for each appended completion with a pre-trained PointNet~\cite{qi2017pointnet}. We use t-SNE~\cite{van2008visualizing} to project each feature representation into a point in~\cref{tsne}. From~\cref{tsne}, we see that 1) the visualized 3D representations completed from a same partial input are clustered, which shows that our learned prior model is able to capture common structural features for the same category; and 2) results from different inputs are obviously scattered, which demonstrates that the learned prior distribution is diverse enough (in other words, the coverage of the learned prior is sufficiently wide), so that it can differentiate different object categories. The above observations confirm that our learned prior is able to well model the underlying dependencies between local shape components, thus it serves as a general object structure probabilistic distribution, supporting a wide range of related tasks.

\begin{table}
 \vspace{-3mm}
  \centering
  \resizebox{\linewidth}{!}
  {
  \begin{tabular}{lcccccc}
  \toprule
               & \multicolumn{3}{c}{Bottom Half}    & \multicolumn{3}{c}{Octant}     \\  \cline{2-7} 
Method       & MMD$\downarrow $  & AMD$\downarrow $        & TMD$\uparrow $        & MMD$\downarrow $  & AMD$\downarrow $     & TMD$\uparrow $       \\
\midrule
PoinTr~\cite{yu2021pointr}     &  0.5316    & N/A    & N/A      &   2.1567       & N/A    & N/A    \\
SeedFormer~\cite{zhou2022seedformer}     & 0.4972  & N/A  & N/A    &2.3990     &N/A           & N/A        \\
AutoSDF~\cite{mittal2022autosdf}    & 0.3510   & 0.8200     & 0.0466  & 0.5720  & 1.279   & 0.0826      \\
\midrule
Ours  & \textbf{0.2933} & \textbf{0.6302} & \textbf{0.0478} & \textbf{0.4690}  & \textbf{1.093}  & \textbf{0.0960}\\

\bottomrule

  \end{tabular}}
  \caption{Quantitative completion results on ShapeNet. MMD and AMD is multiplied by $10^2$. TMD is multiplied by $10$.}
  \label{tab:completion}
   \vspace{-3mm}
\end{table}

\begin{table}
  \centering
  \resizebox{\linewidth}{!}
  {
  \begin{tabular}{lcccc}
  \toprule
Method       & PMMD$\downarrow $        & CLIP-S$\uparrow $        & FPD$\downarrow $     & TMD$\uparrow $       \\
\midrule
Shape-IMLE~\cite{liu2022towards}     & 1.681          & 31.42         & 82.34          & 0.0539        \\
AutoSDF~\cite{mittal2022autosdf}    & 1.961        & 31.65         & 141.87          & 0.1302        \\
\midrule
Ours  & \textbf{1.492} & \textbf{32.11} & \textbf{59.00} & \textbf{0.2795} \\

\bottomrule

  \end{tabular}}
  \caption{Quantitative results for text-guided generation. PMMD and CLIP-S is multiplied by $10^2$. TMD is multiplied by $10$.}
  \label{tab:text}
   \vspace{-5mm}
\end{table}

\subsection{Language-guided Generation}
\label{sec:text}
 \vspace{-2mm}
 
\noindent
\textbf{Data and evaluation.} We reorganize the dataset released by ShapeGlot~\cite{achlioptas2019shapeglot} into text-shape pairs to train a text-driven conditional generative model for 3D shapes, using the train/test splits provided by AutoSDF~\cite{mittal2022autosdf}.
Since most existing metrics cannot well measure the similarity between text and shape modalities, we propose new evaluation metrics for text-driven shape generation task. \textit{CLIP-S} computes the maximum score of cosine similarity between $N=9$ generated shapes and their text prompts by a pre-trained CLIP~\cite{radford2021clip}. Since CLIP cannot handle 3D shape inputs, we render each generated shape into 20 2D images from different views to compute \textit{CLIP-S}. In addition, We deploy \textit{Frechet Pointcloud Distance} (FPD) and \textit{Pairwise Minimum Matching Distance} (PMMD) to calculate the distance between ground truth and samples.

 \begin{figure}[t]
 \vspace{-5mm}
  \centering
   \includegraphics[width=\linewidth]{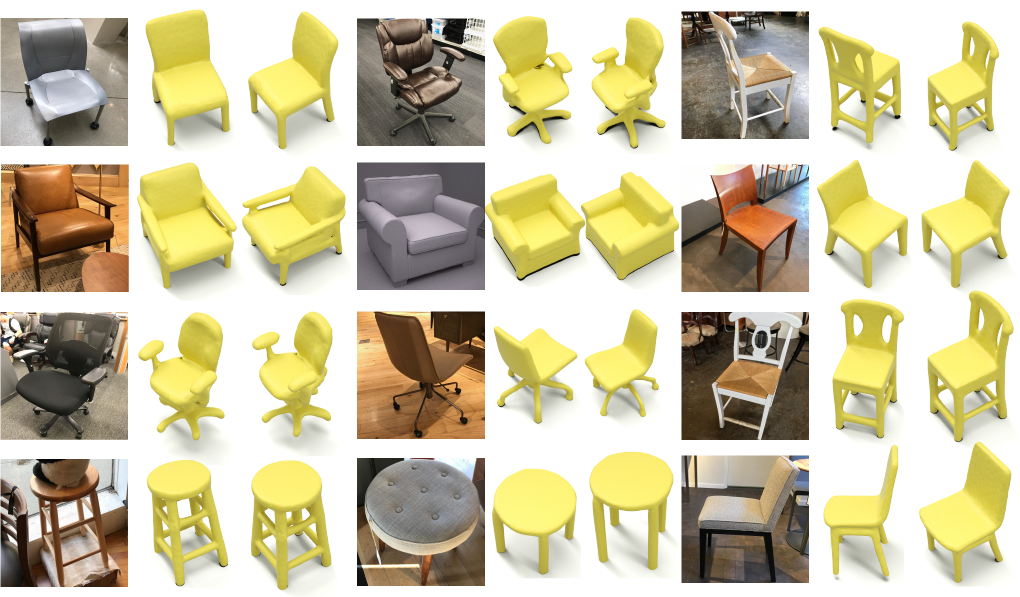}
   \caption{Single-view reconstruction results. Our model synthesizes high-quality single-view reconstruction with learned prior and cross-domain alignment.}
   \label{img_3d}
   \vspace{-5mm}
\end{figure}

\noindent
\textbf{Condition injection.} We first use the CLIP~\cite{radford2021clip} pre-trained model (ViT-B) to encode text prompts to 77 tokens. Then parts of self-attention modules in the denoising network are replaced by cross-attention~\cite{chen2021crossvit}, for its natural compatibility with multi-modality data. As a result, the text token embeddings constantly influence the generation through cross-attention modules.

\noindent
\textbf{Baselines and results.} We quantitatively compare our results with two state-of-the-art methods, \ie, AutoSDF~\cite{mittal2022autosdf} and Shape-IMLE~\cite{liu2022towards}. We only use the generated shapes from Shape-IMLE~\cite{liu2022towards} and discard their colors, to only evaluate 3D geometry quality. Quantitative results are reported in~\cref{tab:text} and a visual comparison is shown in~\cref{text_3d}. We observe a lack of diversity from results of baselines, while our method is diversified due to randomness in diffusing process. The unified discrete local representation of codebook indexing among texts and shapes achieves good alignment between two modalities.

\subsection{Extended Applications}
\label{sec:Extensions}
 \vspace{-2mm}
 
In addition to the above major tasks, we show that 3DQD can be extended for a wide range of downstream applications, serving as a generalized prior \textbf{with little or even no tuning}. Note that in contrast, previous 3D shape priors~\cite{zhou2021pvd, mittal2022autosdf} DO NOT possess this generalization capability.

\noindent
\textbf{Denoising conditonal generation.}
In practice, 3D shapes captured from real scenes often have rough surfaces and noise points due to precision limitation of the capture device. It thus requires extra adaption for downstream models to accommodate noisy data, \eg point clouds denoising~\cite{rakotosaona2020pointcleannet}. To demonstrate our model's ability in dealing with noisy inputs, we add different levels of Gaussian and uniform noise to T-SDFs to simulate the noisy mesh surfaces in real world. Then noisy T-SDFs are encoded into shape tokens as $\widetilde{s}_k$ to replace the fully masked $s_T$ into our pre-trained 3DQD, where $k=\frac{1}{2}T$. Then reverse Markov process starts from $\widetilde{s}_k$. Visual results are shown in~\cref{uncond_denoise}, and quantitative evaluations with different noise levels and types are detailed in our supplementary material.
From~\cref{uncond_denoise}, it is noted that our model successfully recovers noisy inputs into clean samples \textbf{without any tuning}, demonstrating the noise immunity ability of our pre-trained 3DQD, which comes from the noise compatibility from diffusing training and quantized noise-free vectors in codebook.

\begin{figure}[t]
\vspace{-7mm}
  \centering
   \includegraphics[width=\linewidth]{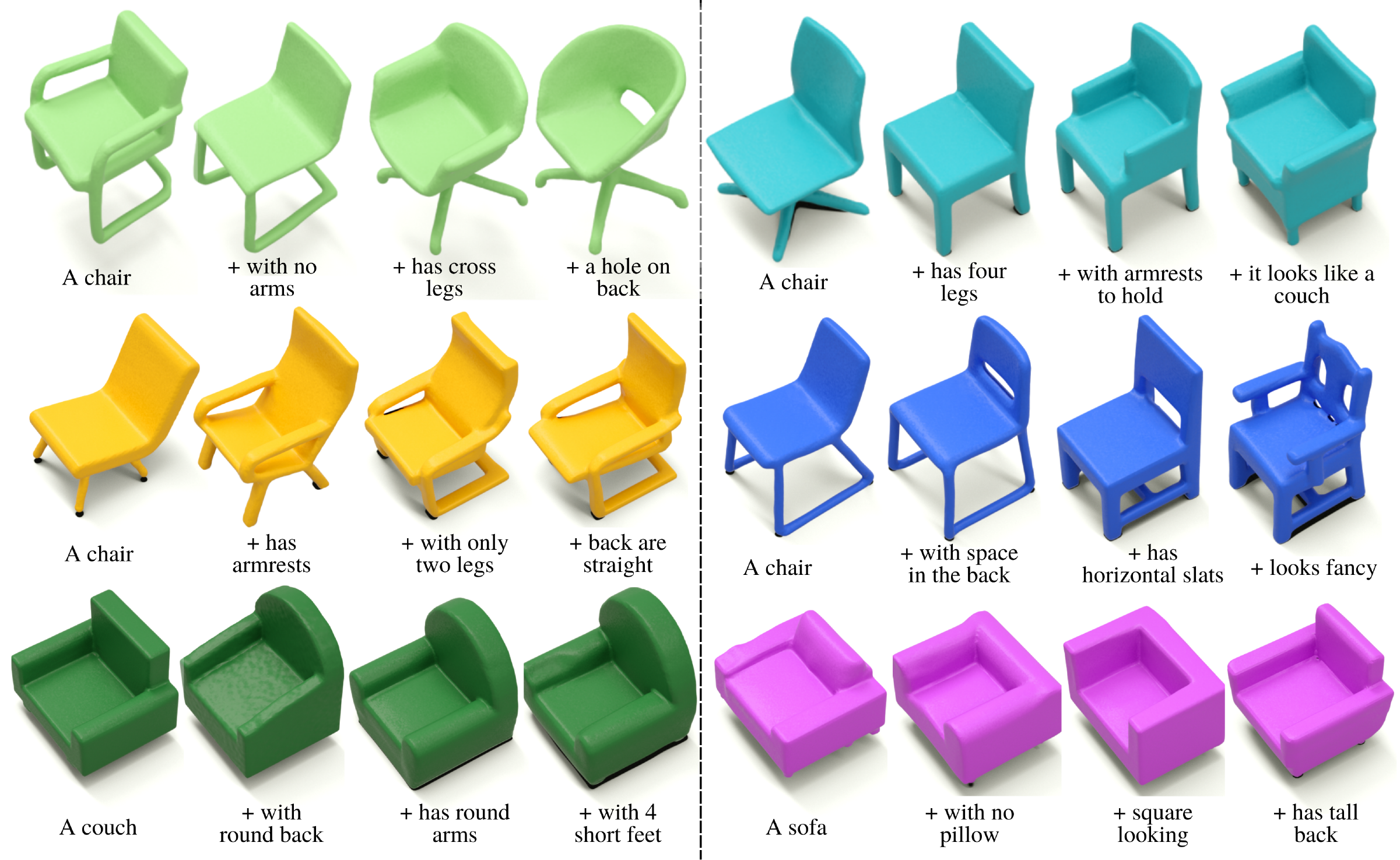}
   \caption{Text-driven shape editing results. We start from initial phrase at left and gradually edit them with new prompts. They present continuous and smooth evolution.}
   \label{text_edit}
    \vspace{-4mm}
\end{figure}

\noindent
\textbf{Text-driven shape editing.}
We also conduct an experiment on text-guided shape editing \textbf{without any tuning}. We initialize standard chair token maps with text prompts \textit{``A chair"} with 3DQD. Then we initialize the former token maps as the start of new reverse process $\widetilde{s}_k$ for sequentially generating new shapes, with new text inputs one by one. $k$ is set to $0.98T$ to encourage novel structure generation. The results are visualized in~\cref{text_edit}. Generated shapes are apparently high-quality and realistic, due to the structure prior memorized by diffusion generator.

\begin{table}
%\vspace{-7mm}
  \centering
  \resizebox{\linewidth}{!}{
  \begin{tabular}{lccc}
  \toprule
Method       & VAE Training       & Diffusion Training        & Inference Time / Shape     \\
\midrule
Lion~\cite{zeng2022lion}     & 110 hours          & 440 hours         & 27.12 seconds                \\
Ours    & \textbf{9 hours}       & \textbf{60 hours}        & \textbf{1.61 seconds}           \\

\bottomrule
%\vspace{-5mm}
  \end{tabular}}
  \caption{Quantitative comparison of computational cost on a single V100 NVIDIA GPU. Our VAE is fine-tuned on pre-trained model released in ~\cite{mittal2022autosdf} for 9 hours.}
  \label{tab:efficiency}
   \vspace{-5mm}
\end{table}

\noindent
\textbf{Single-view reconstruction.}
Inferring the 3D shape from a single image is always non-trivial due to the lack of information. We show that with a well pre-trained VQ-VAE to encode and quantize images, our model achieves singe-view 3D reconstruction with a little fine-tuning. Specifically, we obtain the index coding of images with VQ-VAE released by Mittal \etal~\cite{mittal2022autosdf} on Pix3D dataset~\cite{sun2018pix3d}, and modify the condition embedding module of language-guided 3DQD model from discrete text tokens to discrete image tokens, with rest parts of the net remaining unchanged. We then fine-tune this conditional 3DQD model on Pix3D with masked image-shape pairs for 10 hours. For inference, we set index codes of images from VQ-VAE at the beginning of reverse process $\widetilde{s}_k$. Note that image embeddings modulate the generation process via cross-attention. Visual results are shown in~\cref{img_3d} as accurate and realistic reconstructions with learned prior and cross-domain alignment. More results are provided in our supplementary material.

\begin{figure}[t]
 \vspace{-5mm}
  \centering
   \includegraphics[width=\linewidth]{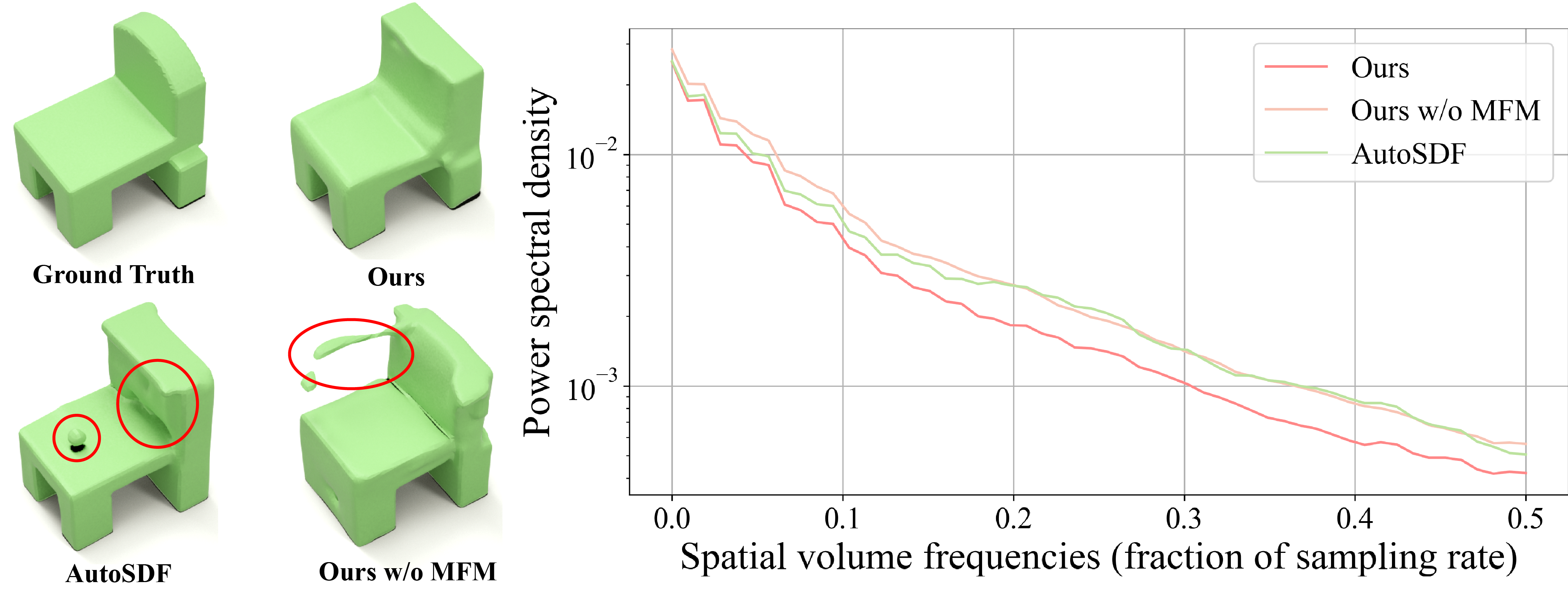}
   \caption{Power spectral density of different frequencies on 1000 generated shapes along with visual comparison. Our MFM layers successfully suppress high-frequency components.}
   \label{freq}
 \vspace{-5mm}
\end{figure}

 \vspace{-2mm}
\subsection{Ablation Study}
\label{sec:Ablation}

 \vspace{-2mm}
\subsubsection{Frequency Analysis}
\label{sec:freq}
 \vspace{-2mm}
We experimentally analyze the frequency components of the generation to validate our design of Multi-frequency Fusion Module (MFM). We sample 1000 completion samples with three methods: our 3DQD, our 3DQD without MFM, and AutoSDF. The T-SDF results are fed to Discrete Cosine Transform to calculate its average power spectral density and obtain frequency components within each spatial volume. The power spectral density of sampled volumes is plotted in~\cref{freq}. It is noted that our MFM layers successfully suppress high-frequency noise and help generate shapes with smooth surfaces. More ablation studies about the architecture of MFM is in our supplementary material.

 \vspace{-3mm}
\subsubsection{Efficiency Analysis}
\label{sec:Efficiency}
 \vspace{-2mm}
 
We compare our model with the latest 3D diffusion work Lion~\cite{zeng2022lion} (with no open source codes) about training and inference complexity, evaluated on single-class shape generation task as shown in~\cref{tab:efficiency}. It is observed that our model only requires 69 (9+60) hours for training totally while Lion~\cite{zeng2022lion} spends as much as 550 (110+440) hours. Also note that our reported results are all generated with 100-step DDPM-based sampling, much more efficient than that of Lion (\ie 1000 steps), with visually competitive performance compared to Lion~\cite{zeng2022lion}.

 \vspace{-3mm}
\section{Conclusion} 
 \vspace{-2mm}

We have introduced 3DQD, a unified and efficient shape generation prior model for multiple 3D tasks. Our model first learns a compact representation with P-VQ-VAE for its advantages in computational saving and consistency among different tasks. Then a novel discrete diffusion generator is trained with accurate, expressive and diversified object structural modeling. Multi-frequency fusion modules are developed to suppress high-frequency outliers. Solid experiments and rich analysis have demonstrated our approach possesses superior generative power and impressive shape generation quality on various 3D shape generation tasks. Furthermore, our prior model can serve as a generalized backbone for multiple downstream tasks with no or little tuning, while no architectural change is needed.

 \vspace{-2mm}
\section{Acknowledge} 
 \vspace{-2mm}
This work was supported by National Science Foundation of China (U20B2072, 61976137). This work was also partly supported by SJTU Medical Engineering Cross Research Grant YG2021ZD18.

%%%%%%%%% REFERENCES
{\small
\bibliographystyle{ieee_fullname}
\bibliography{egbib}
}

\end{document}

\title{Supplementary Materials:   

3DQD: Generalized Deep 3D Shape Prior via Part-Discretized Diffusion Process
}

\author{Yuhan Li\textsuperscript{1}\hspace{2mm}
Yishun Dou\textsuperscript{2}\hspace{2mm}
Xuanhong Chen\textsuperscript{1}\hspace{2mm}
Bingbing Ni\textsuperscript{1, 2$\dagger$}\hspace{2mm}
Yilin Sun\textsuperscript{1}\hspace{2mm}
Yutian Liu\textsuperscript{1}\hspace{2mm}
Fuzhen Wang\textsuperscript{1}\\
\textsuperscript{1}Shanghai Jiao Tong University, Shanghai 200240, China \qquad \textsuperscript{2}Huawei Hisilicon \\
{\tt\small \{melodious, nibingbing\}@sjtu.edu.cn}\\
{\small \url{https://github.com/colorful-liyu/3DQD}}
}

\maketitle

\newcommand{\customfootnotetext}[2]{{% Group to localize change to footnote
  \renewcommand{\thefootnote}{#1}% Update footnote counter representation
  \footnotetext[0]{#2}}}% Print footnote text
\customfootnotetext{${\dagger}$}{Corresponding author: Bingbing Ni.}

%%%%%%%%% BODY TEXT

%-----------------------------------------------------------------------------------

We first provide the implementation details of the P-VQ-VAE, discrete diffusion generator and condition pipeline in~\cref{sec:imple}. More ablation study about important settings is reported in~\cref{sec:more_able}. Technical details about experiments are given in~\cref{sec:expe}, with more visual results in~\cref{sec:visual}.

\section{Implementation} 
\label{sec:imple}
\subsection{P-VQ-VAE Backbone} 

\paragraph{Architecture details.} 
Our P-VQ-VAE backbone consists of three components: an encoder $E$, a decoder $D$ and a vector quantizer $VQ$ with convolutions. Following AutoSDF~\cite{mittal2022autosdf}, we adapt the VQ-VAE from the VAE backbone of LDM~\cite{rombach2021ldm}. We show the details of encoder in \cref{tab:encoder}, the decoder in \cref{tab:decoder}, and the vector quantizer in \cref{tab:vq}. 

\paragraph{Dataset details.} 
We train the P-VQ-VAE using the objects from 13 categories of ShapeNet~\cite{chang2015shapenet} data, including [\textit{airplane, bench, cabinet, car, chair, display, lamp, speaker, rifle, sofa, table, phone, watercraft}]. We first extract the Truncated-SDF (T-SDF) following pre-processing steps in DISN~\cite{xu2019disn} and PixelTransformer~\cite{tulsiani2021pixeltransformer}. The shapes are normalized to lie in an origin-centered cube in $[-1, 1]^3$, while most shape T-SDFs' absolute values are less than $0.5$. The signed distance function is evaluated at locations in a uniformly sampled $64^3$ grid. Following AutoSDF~\cite{mittal2022autosdf}, we use $0.2$ as the threshold to further obtain
the T-SDFs representations.

\paragraph{Training details.} 
Then the whole 3D shape in the format of T-SDF $X \in \mathbb{R }^{64\times 64\times 64}$ is divided into 512 partial regions $X'\in \mathbb{R }^{N\times 8\times 8\times 8}$, and $N=512$ is the number of non-overlap regions, for directly working on whole shape is computationally unaffordable with cubic increase with resolution. Afterward, all regions are vectorized by the encoder as $Z = E(X) \in \mathbb{R }^{N\times n_z}$, while each patch is treated independently and equally.

Next, we obtain a spatial collection of quantized shape tokens $Z_q=\left \{ z_{q}^0, \dots, z_{q}^{N-1} \right \}$ by a further quantization step as $z_{q}^i = VQ(z^{i}) \in \mathbb{R }^{n_z}$, and their corresponding codebook indexes $S=\left \{ s^{0}, \dots, s^{N-1} \right \}$, where $s^i\in \left \{ 0, \dots, K-1 \right \}$. $VQ$ is a spatial-wise quantizer which maps each partial shape $z^i$ into its closest codebook entry in $\mathcal{Z} $. Complete latent space $Z$ and $Z_q$ is produced by gathering $z^i$ and $z_q^i$ for all location $i$ into grids. Finally, decoder $D$ decodes quantized feature $Z_q$ to output the reconstruction $X'$. The training objective consists of three parts: reconstruction loss, VQ loss and commitment loss:

\begin{equation}
L=\underset{Reconstruction~loss}{ \underbrace{\log P(X|Z_q)}} + \underset{VQ~loss}{ \underbrace{\left \| sg\left [ Z \right ] -Z_q \right \| ^2}} + \beta \underset{Commitment~loss}{ \underbrace{\left \|  Z -sg\left [ Z_q \right ]  \right \| ^2}},
  \label{vqvaeloss}
\end{equation}
where $sg[Z]$ means stopping the gradient of Z, so the optimization will only impact the other item. The learning rate is set to be $1e-4$, which halves every 30 epochs. Adam optimizer is used with betas = [0.5, 0.9].

\begin{table*}[]
  \centering{
\begin{tabular}{lccc}
  \toprule
$E$ Layer name & Parameters          & Input size                            & Output size \\
\midrule
Conv3D     & kernel size 3  & $Bs\times1\times8\times8\times8$ & $Bs\times64\times8\times8\times8$            \\
3D ResNet Block  & kernel size 3  & $Bs\times64\times8\times8\times8$   & $Bs\times64\times8\times8\times8$    \\
Down-sample  & kernel size 3, stride 2, padding 0 & $Bs\times64\times8\times8\times8$   & $Bs\times64\times4\times4\times4$    \\    
3D ResNet Block  & kernel size 3  & $Bs\times64\times4\times4\times4$   & $Bs\times128\times4\times4\times4$    \\
Down-sample  & kernel size 3, stride 2, padding 0 & $Bs\times128\times4\times4\times4$   & $Bs\times128\times2\times2\times2$    \\  
3D ResNet Block  & kernel size 3  & $Bs\times128\times2\times2\times2$   & $Bs\times128\times2\times2\times2$    \\
Down-sample  & kernel size 3, stride 2, padding 0 & $Bs\times128\times2\times2\times2$   & $Bs\times128\times1\times1\times1$    \\  
3D ResNet Block  & kernel size 3  & $Bs\times128\times1\times1\times1$   & $Bs\times256\times1\times1\times1$    \\
3D ResNet Block  & kernel size 3  & $Bs\times256\times1\times1\times1$   & $Bs\times256\times1\times1\times1$    \\
3D Attention Block  & Multi-head 1 & $Bs\times256\times1\times1\times1$   & $Bs\times256\times1\times1\times1$    \\
3D ResNet Block  & kernel size 3  & $Bs\times256\times1\times1\times1$   & $Bs\times256\times1\times1\times1$    \\
Group Norm   & number of groups 32   &    $Bs\times256\times1\times1\times1$   & $Bs\times256\times1\times1\times1$     \\
Swish     &       -       &    $Bs\times256\times1\times1\times1$     &  $Bs\times256\times1\times1\times1$    \\
Conv3D    &kernel size 3    &  $Bs\times256\times1\times1\times1$     &  $Bs\times256\times1\times1\times1$      \\
\bottomrule
\end{tabular}}
\caption{Architecture of encoder $E$ of P-VQ-VAE. The default stride is set to 1.  The default padding is set to 1.}
  \label{tab:encoder}
\end{table*}

\begin{table*}[]
  \centering{
\begin{tabular}{lccc}
  \toprule
$VQ$ Layer name & Parameters          & Input size                            & Output size \\
\midrule
Conv3D     & kernel size 3  & $Bs\times256\times1\times1\times1$ & $Bs\times256\times1\times1\times1$  \\Quantizer     & codebook $K \times 256$  & $Bs\times256\times1\times1\times1$ & $Bs\times256\times1\times1\times1$  \\
Conv3D    &kernel size 3 & $Bs\times256\times1\times1\times1$  &$Bs\times256\times1\times1\times1$  \\
\bottomrule
\end{tabular}}
\caption{Architecture of vector quantizer $VQ$ of P-VQ-VAE. The default stride is set to 1.  The default padding is set to 1.}
  \label{tab:vq}
\end{table*}

\begin{table*}[]
  \centering{
\begin{tabular}{lccc}
  \toprule
$D$ Layer name & Parameters          & Input size                            & Output size \\
\midrule
Conv3D     & kernel size 3  & $Bs\times256\times1\times1\times1$ & $Bs\times256\times1\times1\times1$            \\
3D ResNet Block  & kernel size 3  & $Bs\times256\times1\times1\times1$   & $Bs\times256\times1\times1\times1$    \\
3D Attention Block  & Multi-head 1 & $Bs\times256\times1\times1\times1$   & $Bs\times256\times1\times1\times1$    \\
3D ResNet Block  & kernel size 3  & $Bs\times256\times1\times1\times1$   & $Bs\times256\times1\times1\times1$    \\
Up-sample  & kernel size 3, stride 2, padding 0 & $Bs\times256\times1\times1\times1$   & $Bs\times256\times2\times2\times2$    \\  
3D ResNet Block  & kernel size 3  & $Bs\times256\times2\times2\times2$   & $Bs\times128\times2\times2\times2$    \\
3D Attention Block  & Multi-head 1 & $Bs\times128\times2\times2\times2$   & $Bs\times128\times2\times2\times2$    \\
Up-sample  & kernel size 3, stride 2, padding 0 & $Bs\times128\times2\times2\times2$   & $Bs\times128\times4\times4\times4$    \\  
3D ResNet Block  & kernel size 3  & $Bs\times128\times4\times4\times4$   & $Bs\times64\times4\times4\times4$    \\
Up-sample  & kernel size 3, stride 2, padding 0 & $Bs\times64\times4\times4\times4$   & $Bs\times64\times8\times8\times8$    \\  
3D ResNet Block  & kernel size 3  & $Bs\times64\times8\times8\times8$   & $Bs\times64\times8\times8\times8$     \\
3D ResNet Block  & kernel size 3  & $Bs\times64\times8\times8\times8$   & $Bs\times64\times8\times8\times8$     \\
Group Norm   & number of groups 32   &    $Bs\times64\times8\times8\times8$   & $Bs\times64\times8\times8\times8$     \\
Swish     &       -       &    $Bs\times64\times8\times8\times8$     &   $Bs\times64\times8\times8\times8$    \\
Conv3D    &kernel size 3    &  $Bs\times64\times8\times8\times8$     &  $Bs\times1\times8\times8\times8$     \\
\bottomrule
\end{tabular}}
\caption{Architecture of decoder $D$ of P-VQ-VAE. The default stride is set to 1.  The default padding is set to 1.}
  \label{tab:decoder}
\end{table*}

%-----------------------------------------------------------------------------------

\subsection{Discrete Diffusion Generator}

\begin{figure*}[t]
  \centering
   \includegraphics[width=0.9\linewidth]{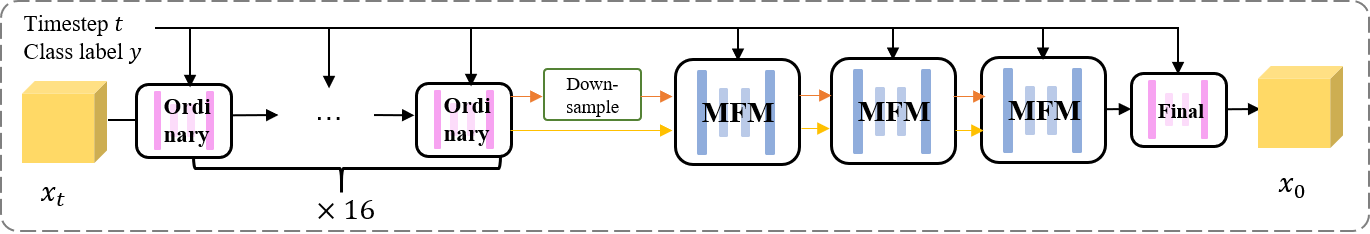}
   \caption{The complete denoising transformer framework which is used in reverse process.}
   \label{diff_pipeline}
\end{figure*}

\paragraph{Architecture details.} 
We adopt a transformer-based architecture to model the joint probabilistic distribution prior over 3D shapes. The transformer consists of 16 Ordinary Blocks, 3 Multi-frequency Modules (MFM), and a Final Block with the same setting as Ordinary Blocks, as shown in~\cref{diff_pipeline}.

\begin{figure*}[t]
  \centering
   \includegraphics[width=0.8\linewidth]{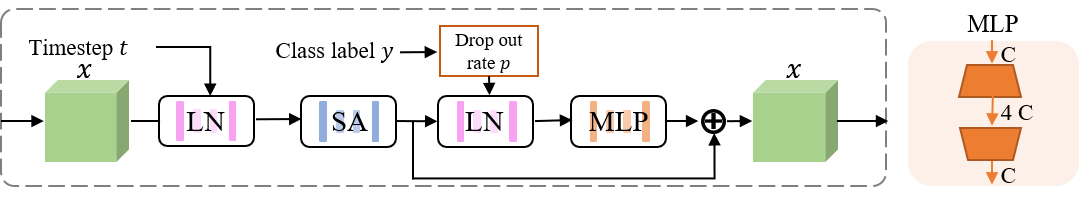}
   \caption{Ordinary block with a boosting bottleneck MLP.}
   \label{ordinary}
\end{figure*}

We show the details of the Ordinary Block in \cref{ordinary}. The default dimension of latent feature is $C=256$. All activation function is \textit{GELU2()}. AdaLayerNorm layers are used to synthesize timestep and class-label information and modulate the whole generation process, since timestep $t$ is an important parameter to control the degree of denoising of outputs, while MLP has a boosting bottleneck of $4\times C$ channels. The MFM doubles the branches of information flowing in Ordinary Blocks with intra- and inter-communication. 

\begin{figure*}[t]
  \centering
   \includegraphics[width=0.8\linewidth]{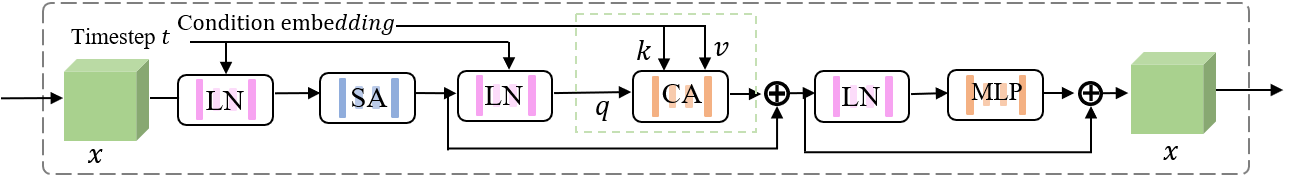}
   \caption{Ordinary block with cross-attention to modulate multi-modality information.}
   \label{ca_ordinary}
\end{figure*}

\paragraph{Training details.} 
After the traning of P-VQ-VAE, each shape is encoded into quantized shape tokens $Z_q$ with their codebook index map $S=\left \{ s^{0}, \dots, s^{N-1} \right \}$, where $s^i\in \left \{ 0, \dots, K-1 \right \}$. The index map is sent to the forward diffusing process followed by reverse learnable denoising chain. In this stage, the encoder and the decoder are frozen, and only the diffusion generator network is trained using the objectives with uniformly sampled timestep $t$ from $\left \{ 1, 2, \dots, 100 \right \} $:
\begin{equation}
L=-\mathbb{E} _{q(s_0)q(s_t|s_0)}\left [\underset{L_{main}}{\underbrace{\log p_{\theta}(s_{t-1}|s_t)}} +\underset{L_{aux}}{\lambda \underbrace{\log p_{\theta}(\hat{s}_0|s_t)}} \right ],
  \label{Loss}
  \vspace{-2mm}
\end{equation}
where $\lambda $ is $1e-3$.

Following minGPT~\cite{radford2019language}, we separate out all parameters of the model into two buckets: some will experience weight decay for regularization and the others won't (\ie biases, and layernorm / embedding weights). The learning rate is set to be $1e-4$, which multiplies 0.9 every 30 epochs. Besides, to stabilize the training, we convert one-hot row vector $s$ to log one-hot vector with a minimum value of $-69.07$ ($ln(-30)$) according to Argmax Flow~\cite{hoogeboom2021argmax}.

%-----------------------------------------------------------------------------------

\subsection{Condition Injection Module} 
\label{sec:condition}
In face of the challenge of various conditional generation tasks like
shape completion, or generation based on different modalities like images and language, considering a single way to solve all problems is not possible. We design two ways to inject the different modality information which are concisely introduced in main body of this paper. More details are given here.

\paragraph{Shape initialization as conditions.}
Generally speaking, in the inference stage, the diffusion generator starts the reverse denoising process from completely corrupted or masked shape tokens $S_T$ and recovers the token maps $S_T$ towards the desired data distribution $S_0$. In this case, the token maps $S_T$ are from the prior setting (\ie, fully masked matrices). However, if the shapes or parts of shapes $\tilde{s}_0$ (\eg the editing task and completion task with given shapes) are provided, we will go on the forward diffusing process for $k$ timesteps on $\tilde{s}_0$ and obtain partially corrupted conditional feature maps $\tilde{s}_k$, where some original shape structure information still remains. Initialized with $\tilde{s}_k$ rather than fully masked tokens $S_T$, the learned reverse process can be derived as:
\begin{equation}
p_{\theta }(\tilde{s}_{0:k})=p(\tilde{s}_k) {\textstyle \prod_{t=1}^{k}p_{\theta}(\tilde{s}_{t-1}|\tilde{s}_t)},
  \label{cond1}
\end{equation}
gradually removing the added noise and generating a 3D shape from condition. Smaller $k$ value means corrupting the condition sightly, so the freedom to denoise and reorganize conditionally initialized token maps is quite restricted, leaning towards less diversity and limited rebuilding capability (\ie, worse TMD and MMD). While a large $k$ value means hurting the condition seriously, it always leads to low-fidelity and distorted generated results, violating the condition (\ie, bad MMD).

We use this kind of condition injection pipeline for shape completion task ($k=0.5\times T=50$), shape denoising task ($k=0.5\times T=50$), and text-driven shape editing ($k=0.98\times T=98$).

\paragraph{Cross-attention for conditions.} Although shape initialization solves some problems, not all conditions can be expressed as token maps initialization, especially condition $c$ with different modalities (\eg, images, texts). As a consequence, inspired by ~\cite{rombach2021ldm}, we use cross-attention modules to inject conditions. At first, conditions with different modalities are encoded into discrete and quantized tokens embedding with its own encoder. For texts, we utilize pre-trained CLIP~\cite{radford2021clip} model (ViT-B) to project text prompts to 77 tokens $c_q$. As for single-view images, VQ-VAE from ~\cite{mittal2022autosdf} is used to encode images to 512 tokens $c_q$, which is specifically trained on image data on Pix3D dataset~\cite{sun2018pix3d} and aligned with shape tokens. Next, only the indexes $c_I$ of these tokens among codebook entries are preserved, while $c_q$ is discarded. The wide use of indexes is beneficial to allow the model to receive uniform input (\ie, conditions and shapes are all expressed with integers), and eliminate the inherent differences in data distribution among different modalities. Then the conditional token indexes are projected to tensors with learnable embedding modules, followed by fully connected layers to encode them to $key$ and $value$ into cross-attention. The Ordinary Block and MFM are modified with a LayerNorm and a cross-attention module behind self-attention, as shown in \cref{ca_ordinary}.

%-----------------------------------------------------------------------------------
%-----------------------------------------------------------------------------------

\section{More Ablation Study} 
\label{sec:more_able}

\subsection{Shape novelty analysis.} 
\begin{figure}[t]
\vspace{-5mm}
  \centering
   \includegraphics[width=\linewidth]{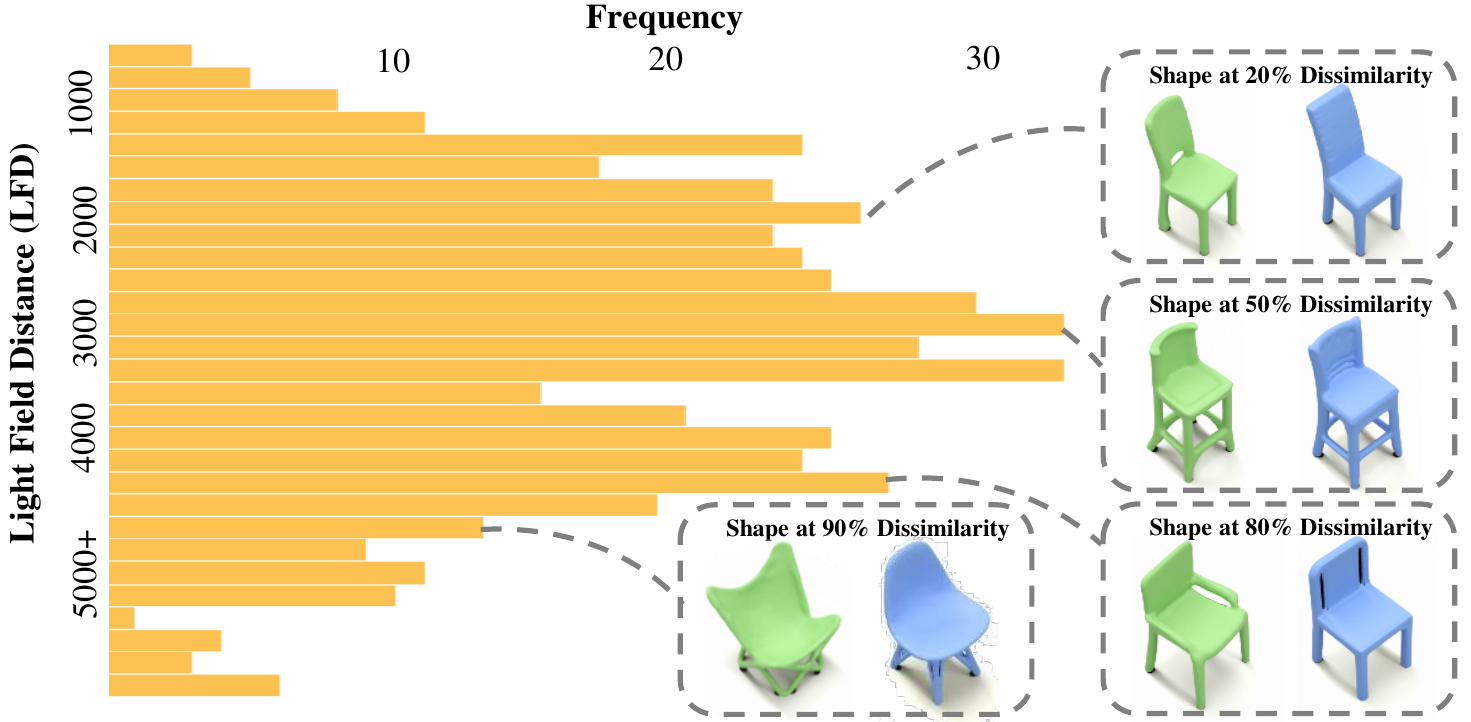}
   \vspace{-3mm}
   \caption{We generate 500 random chairs using 3DQD, then retrieve the most similar shape in training set for each samples with LFD metric. The distribution of LFDs between samples (in green) and retrieved shapes (in blue) is plotted here.}
   \label{lfd}
\vspace{-6mm}
\end{figure}

To statistically analyze the novelty of the generated shapes, we visualize the distribution of LFDs between our generated shapes and retrieved shapes in \cref{lfd}. Note that our 3DQD can not only learn the distribution of training data (low LFD), but also generate novel shapes (high LFD) that are different from the training data.

%-----------------------------------------------------------------------------------
\subsection{Multi-frequency Fusion Module Settings.} 
In the main body of this paper, we have analyzed the frequency components of the generation to validate our design of Multi-frequency Fusion Module (MFM) with quantitative and visual comparisons. In this section, to verify our designed MFM settings, we compare 4 configurations about MFM on the shape completion task:
\begin{itemize}
\item 3DQD model with different MFM layers. It still utilizes residual add as Pair-wise Fusion Operator (PFO) with the complete MFM pipeline.
\end{itemize}

\begin{itemize}
\item 3DQD model with cross-attention to fuse two branches of frequency components. In this case, adaptive rescaling is no longer required, for cross-attention is able to align features with different spatial dimensions with $QKV$.
It implements 3 complete MFM layers.
\end{itemize}

\begin{figure}[t]
  \centering
   \includegraphics[width=\linewidth]{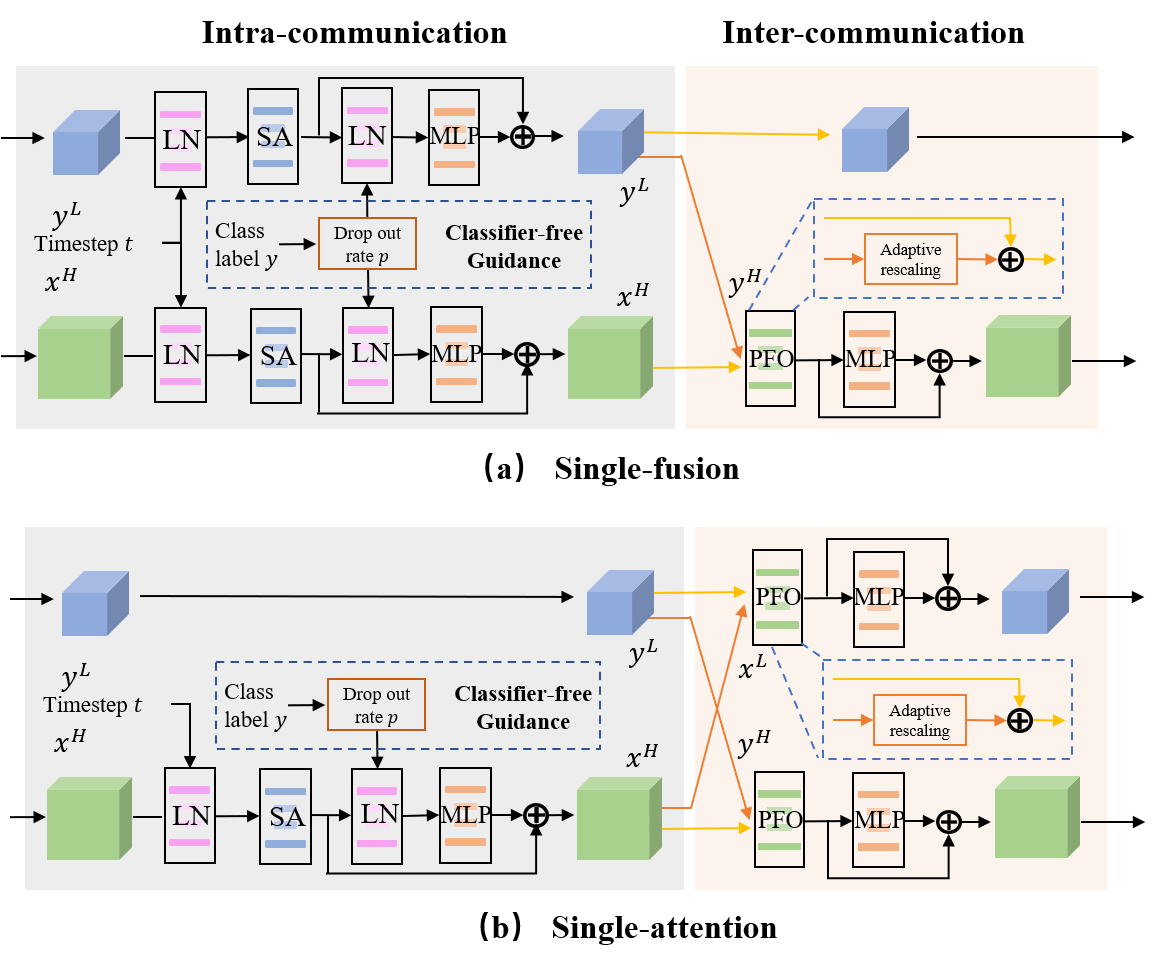}
   \caption{Two MFM communication frameworks in ablation cases (single-fusion and single-attention).}
   \label{Fig: MFM_abla}
\end{figure}

\begin{itemize}
\item 3DQD model with single-fusion MFM layers as shown in \cref{Fig: MFM_abla}. In this case, the inter-communication between $y^L$ and $x^L$ is absent, so the fusion only occurs in the other side. It uses residual add as fusion operator with 3 MFM layers.
\end{itemize}

\begin{itemize}
\item 3DQD model with single-attention MFM layers as shown in \cref{Fig: MFM_abla}. In this case, the intra-communication within $y^L$ is absent. It uses residual add as fusion operator with 3 MFM layers.
\end{itemize}

Quantitative results are shown in \cref{tab:abla_MFM}, where we compare our full MFM pipeline with various ablated cases. The best choice for MFM is utilizing residual add as PFO, with 3 full MFM pipeline layers.

\begin{table}
\resizebox{\linewidth}{!}{
  \centering
  %\resizebox{\linewidth}{!}
  {
  \begin{tabular}{ccccc}
  \toprule
MFM layers & Fusion types & Dual fusion & Dual-attn & MMD$\downarrow $   \\
\midrule
3          & residual add & \ding{52}   & \ding{52} & \textbf{0.2934} \\ \hline \hline

0          &              &             &           & 0.3205 \\
1          &              &             &           & 0.3083 \\
2          &              &             &           & 0.3001 \\
4          &              &             &           & 0.2942 \\
5          &              &             &           & 0.2937 \\ 
           & cross-attn   &             &           & 0.4173 \\
           &              & \ding{56}   &           & 0.2981 \\
           &              &             & \ding{56}   & 0.3059 \\

\bottomrule

  \end{tabular}}}
  \caption{Comparing our full MFM pipeline (the first row) with various ablated cases on the shape completion task. We use the hyperparameters on the first line as default hyperparameters for blanks. MMD is multiplied by $1\times 10^{2}$.}
  \label{tab:abla_MFM}
\end{table}

%-----------------------------------------------------------------------------------
%-----------------------------------------------------------------------------------

\section{Experiment Detail} 
\label{sec:expe}

\subsection{Details for Unconditional Generation} 
\paragraph{Evaluation metrics details.}
Although the outputs of our model are T-SDFs, we choose to evaluate them by sampling point clouds on their surfaces. The community has proposed different metrics to quantitatively evaluate the generation performance of point cloud generative models, but some of them suffer from certain drawbacks. Given a generated set of point clouds $S_g$ and a reference set $S_r$, the most popular metrics are (following Lion~\cite{zeng2022lion}):

\begin{itemize}
    \item \textbf{Coverage(COV)}:
\begin{equation}
COV(S_g, S_r)=\frac{\left | \left \{ \underset{Y\in S_r}{\text{arg min}}D(X,Y)|X\in S_g \right \}  \right | }{\left | S_r \right | } ,
\end{equation}
where $D(\cdot ,\cdot)$ is distance measurement (\ie Chamfer distance or earth mover distance). COV measures the number of reference point clouds that are matched to at least one generated shape. COV can quantify diversity, but low-quality and diverse generated point clouds can achieve high coverage scores.
\end{itemize}

\begin{itemize}
    \item \textbf{Minimum matching distance(MMD)}:
\begin{equation}
MMD(S_g, S_r)=\frac{1}{|S_r|}\sum_{Y\in S_r}\underset{X\in S_g}{\text{min}}D(X,Y). 
\end{equation}
MMD calculates the average distance between the point clouds in the reference set and their closest neighbors in the generated set. However, MMD is not sensitive to low-quality point clouds which will never be matched during calculation. Moreover, MMD varies depending on implementation and is not robust on large scale unconditional shape generation.
\end{itemize}

\begin{itemize}
    \item \textbf{1-nearest neighbor accuracy (1-NNA)}: To overcome the drawbacks of COV and MMD, PointFlow~\cite{yang2019pointflow} proposes to use 1-NNA as a metric to evaluate point cloud generative models:

\begin{equation}
\begin{split}
&1\text{-}NNA(S_g, S_r)=\\
&~~~~~~~~~~~~\frac{\sum_{X\in S_g}\mathbb{I}\left [ N_X\in S_g \right ] + \sum_{Y\in S_r}\mathbb{I}\left [ N_Y\in S_r \right ]}{|S_r|+|S_g|},
\end{split}
\label{1nn}
\end{equation}
where $\mathbb{I}[\cdot ] $ is the indicator function and $N_X$ is the nearest neighbor of X in the set $S_r\cup  S_g- \left \{  X  \right  \}$. Hence, 1-NNA represents the leave-one-out accuracy of the 1-NN classifier defined in \cref{1nn}. More specifically, this 1-NN classifier classifies each sample as belonging to either $S_r$ or $S_g$ based on its nearest neighbor sample within $N_X$. As a consequence, 1-NNA directly quantifies distribution similarity between $S_r$ and $S_g$ and measures both quality and diversity.
\end{itemize}

In conclusion, we are following PointFlow~\cite{yang2019pointflow}, PVD~\cite{zhou2021pvd} and use 1-NNA as our main evaluation metric to quantify shape generation performance. with both CD and EMD distances.

\paragraph{More quantitative results.}
More quantitative results including COV and MMD are reported in~\cref{tab:full_uncond} for an additional comparison, although they may be unreliable metrics for unconditional generation quality. 3DQD outperforms all baselines in experiment for the more reasonable 1-NNA metric.

\paragraph{About normalization method.}
Almost all of the baselines in~\cref{tab:full_uncond} preprocess meshes into point clouds following PointFlow, while we preprocess them into T-SDF following DISN~\cite{xu2019disn}, which causes \textbf{different scales of shapes} in two kinds of datasets. Moreover, the most popular method (PointFlow~\cite{yang2019pointflow}, PVD~\cite{zhou2021pvd}, Lion~\cite{zeng2022lion}, \etc) count the mean and variance of the shapes in the training set, and apply \textbf{dataset-wise normalization} to keep the same scales with the training set during evaluation. For a fair comparison on MMD, dataset-wise normalization is also performed on our generated shapes and our test sets to keep the same scales with the training set of PointFlow~\cite{yang2019pointflow} and PVD~\cite{zhou2021pvd}. There is still a gap on MMD to some extent, which mainly comes from the loss caused by multiple sampling during data preprocessing and evaluation. So we strongly recommend referring to relative distance metrics, such as 1-NNA metric.

\begin{figure}[t]
  \centering
   \includegraphics[width=\linewidth]{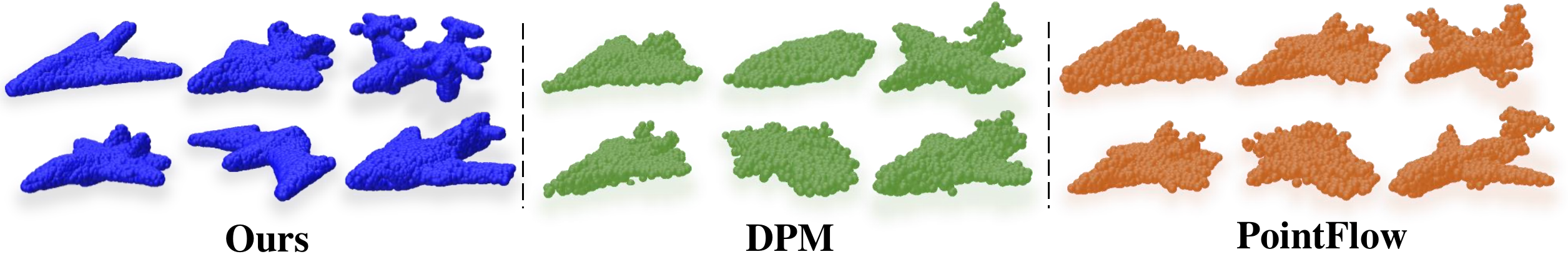}
   
   \caption{We sample shapes with 3DQD model, then we select samples generated by baselines with minimum CD from ours for a fair comparison. Our method outperforms baselines significantly in special shape cases in \textit{Airplane} due to exact shape tokens.}
   \label{air_rebut}
\end{figure}

\paragraph{Improvement on airplane.}
The 20\% performance improvement mainly comes from a large number of special shapes in \textit{airplane} category, such as flat fighters and planes with complex propellers. Our 3DQD generates shapes with exact and clean details in \cref{air_rebut}, because of sufficient and accurate local shape tokens stored in codebook, while previous works are highly possible to produce failure cases with low-quality shape details.

\input{cvpr2023-author_kit-v1_1-1/table/full_uncond}

%-------------------------------------------------------------------------

%-------------------------------------------------------------------------

%-----------------------------------------------------------------------------------
\subsection{Details for Shape Completion} 

\paragraph{Evaluation setting details.}
For PoinTr~\cite{yu2021pointr} and SeedFormer~\cite{zhou2022seedformer}, we use the pre-trained model released on ShapeNet-55 benchmark, which fully covers the 13 categories we used to evaluate. Following PoinTr~\cite{yu2021pointr}, for each object, we randomly sample 8,192 points from the surface to obtain the ground truth point cloud. During evaluation, the points clouds in partial regions (\ie, half and octant) are preserved, while others are discarded. Then the amount of points preserved is repeated or deleted to 4096 (moderate difficulty degree in their benchmark~\cite{yu2021pointr}) as inputs. The models' outputs are 8192 complete points. As for AutoSDF~\cite{mittal2022autosdf} and 3DQD, the partial T-SDFs regions are inputs, and 8192 points are sampled on surfaces of generated shapes. Before computing metrics, all point clouds are normalized with method in SeedFormer~\cite{zhou2022seedformer}.

\paragraph{Evaluation metrics details.}
\textit{Unidirectional Hausdorff Distance} (UHD) is used to measure the completion fidelity to
the input partial input. MPC~\cite{wu2020mpc} and AutoSDF~\cite{mittal2022autosdf} calculate the average Hausdorff distance from the input
partial shape sets $S _p$ to each of the $k$ completion results in $S_g$:
\begin{equation}
UHD(S _p, S_g)=\frac{1}{|S _p|} \sum _{Z\in S_p, X\in S_g}\left ( \frac{1}{k}\sum_{i=1}^k D^{HD}(X^{i}, Z) \right ).
\label{UHD}
\end{equation}
However, UHD only measures the distortion of generated shapes from partial inputs, rather than the quality of completion with ground truths. Low-quality shapes with inputs unchanged will achieve good scores. As a result, we use MMD and Average Matching Distance (AMD) to evaluate completion quality. MMD and AMD for diverse shape completion can be derived as:
\begin{equation}
\begin{split}
& MMD(S_g, S_r)=\frac{1}{|S_r|}\sum_{Y\in S_r, X\in S_g} \left \{ \underset{X_i\in X}{\text{min}}D(X^i,Y)\right \} .\\ 
& AMD(S_g, S_r)=\frac{1}{|S_r|}\sum_{Y\in S_r, X\in S_g}\left \{ \frac{1}{|X|}\sum_{i=0}^{|X|}D(X^i,Y)\right \} . 
\end{split}
\label{MMDAMD}
\end{equation}

%-----------------------------------------------------------------------------------
\subsection{Details for Language-guided Generation} 
Training and inference details are reported in \cref{sec:condition}. As for metrics, \textit{Pairwise Minimum Matching Distance} (PMMD) has the same expression as MMD in~\cref{MMDAMD}. \textit{Frechet Pointcloud Distance} (FPD) is first proposed in TreeGAN to calculate the distance between features of point clouds with a pre-trained PointNet backbone. We select the shapes with minimum PMMD from ground truths in each $k$ generation to build a group of samples, and then measure FPD between the generated data distribution and ground truth distribution on test set, to check if the model can simulate the real shape distribution. Before computing metrics, all point clouds are normalized with method in SeedFormer~\cite{zhou2022seedformer}.

%-----------------------------------------------------------------------------------
\subsection{Details Denoising Conditional Generation} 
The pre-trained single-class 3DQD model trained on \textit{chair} category of ShapeNet data~\cite{chang2015shapenet} is evaluated about its denoising capability in this section to validate the prior on extended applications. We first sample 500 pure T-SDFs inputs $X \in \mathbb{R }^{64\times 64\times 64}$ from the dataset. Then different types of noise $\epsilon \in \mathbb{R }^{64\times 64\times 64}$ (\ie standard Gaussian distribution and uniform distribution) with same spatial resolutions scaled by noise level $\alpha $ are added to pure inputs:
\begin{equation}
X^{noisy} = X + \alpha \epsilon .
\end{equation}
Afterward, the noisy input $X^{noisy}$ is encoded by P-VQ-VAE and set as shape token initialization $\tilde{s}_k$ following the same procedure in \textit{Shape initialization as conditions} part. $k=0.5\times T=50$. At last reverse process starts from $\tilde{s}_k$ to recover the clean samples and remove the noise $\epsilon $.

Quantitative results with different noise levels and types are reported in \cref{tab:denoise}, and visual results are in \cref{sec:visual}.

\input{cvpr2023-author_kit-v1_1-1/table/denoise_tab}
%-----------------------------------------------------------------------------------
\subsection{Details for Single-view Reconstruction} 

\paragraph{Dataset details.}
We evaluate our proposed method on the real-world benchmark Pix3D~\cite{sun2018pix3d}, using cropped and segmented images as inputs for reconstruction (\ie, the background is removed). Since official train/test splits are only provided for the chair category, we test single-view reconstruction on chair category.

\paragraph{Architecture details.}
AutoSDF~\cite{mittal2022autosdf} has released a VQ-VAE to encode images to 512 tokens $c_q$, which is aligned with shape tokens. Based on it, we finish single-view reconstruction with pre-trained 3DQD model trained on language-guided generation task as the backbone, for it has cross-attention modules to fuse cross-modality information. Almost all parts of pre-trained model are preserved, except the conditional embedding modules. 

\paragraph{Training and inference details.}
During reconstruction, we use two ways of condition injection (\ie, shape token initialization and  cross-attention). Namely, we first encode each image into 512 token indexes $s_0$ with VQ-VAE. Then the token indexes $s_0$ after $k$ timestep corruption are initialized as the start of reverse denoising chain $\tilde{s}_k$, where $k=0.5\times T=50$. Afterward, the image indexes $s_0$ are also injected into 3DQD with new learnable embedding layers and cross-attention modules, which is the same as text-driven shape generation. Note that this image-condition 3DQD model is fine-tuned on Pix3D~\cite{sun2018pix3d} with masked and cropped image-shape pairs for 10 hours, to train the new image embedding and refine the whole model well.

%-----------------------------------------------------------------------------------
%-----------------------------------------------------------------------------------
\section{Additional Experimental Results} 
\label{sec:visual}
\begin{figure*}[]
  \centering
   \begin{center}
   \includegraphics[width=0.9\linewidth]{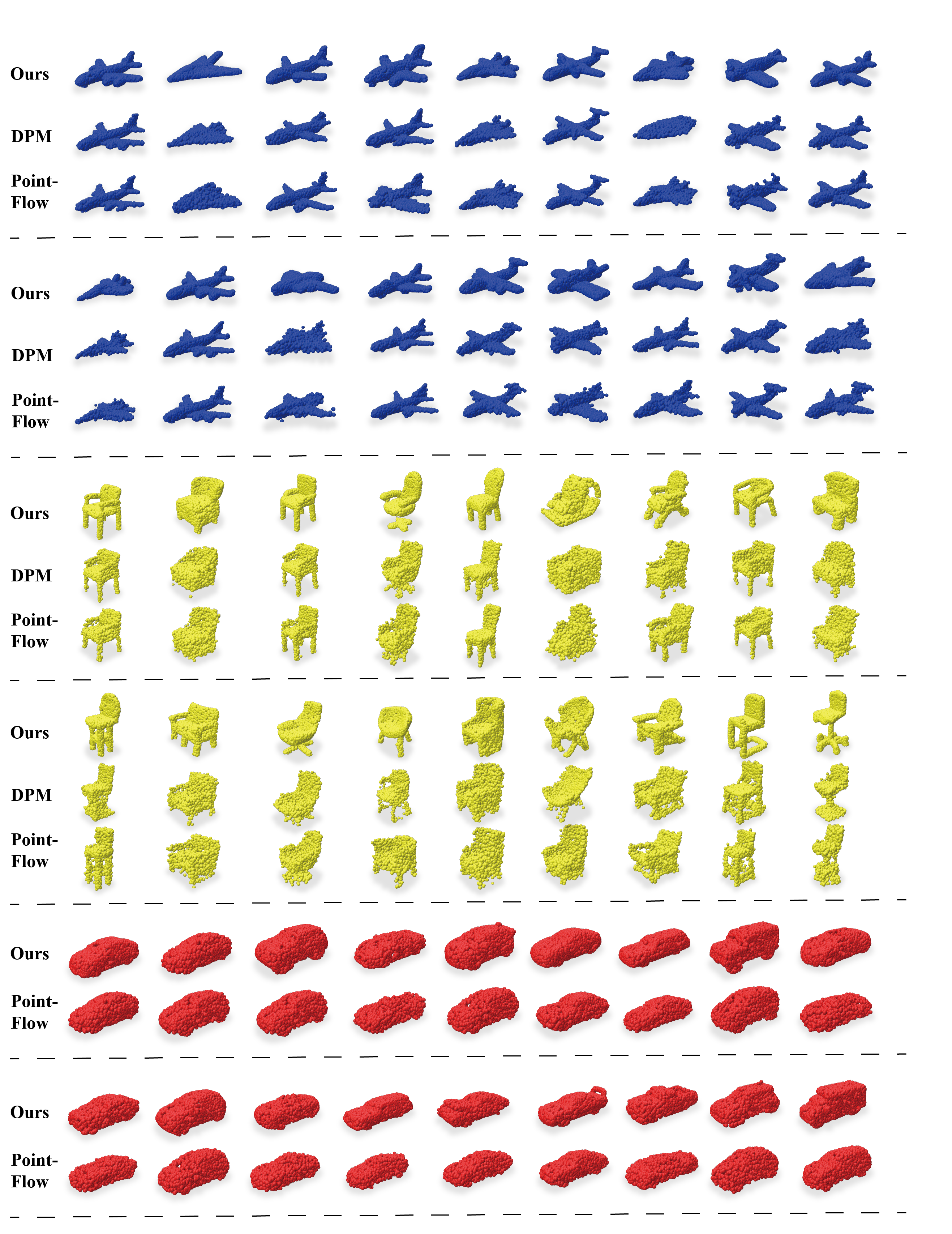}
   \end{center}
   \caption{Visual comparison about unconditional shape generation. We randomly sample shapes with our 3DQD model, then we select samples generated by baselines with minimum Chamfer distances from ours, to present a fair comparison.}
   \label{Fig: supp_uncond}
\end{figure*}

\begin{figure*}[]
  \centering
   \includegraphics[width=0.9\linewidth]{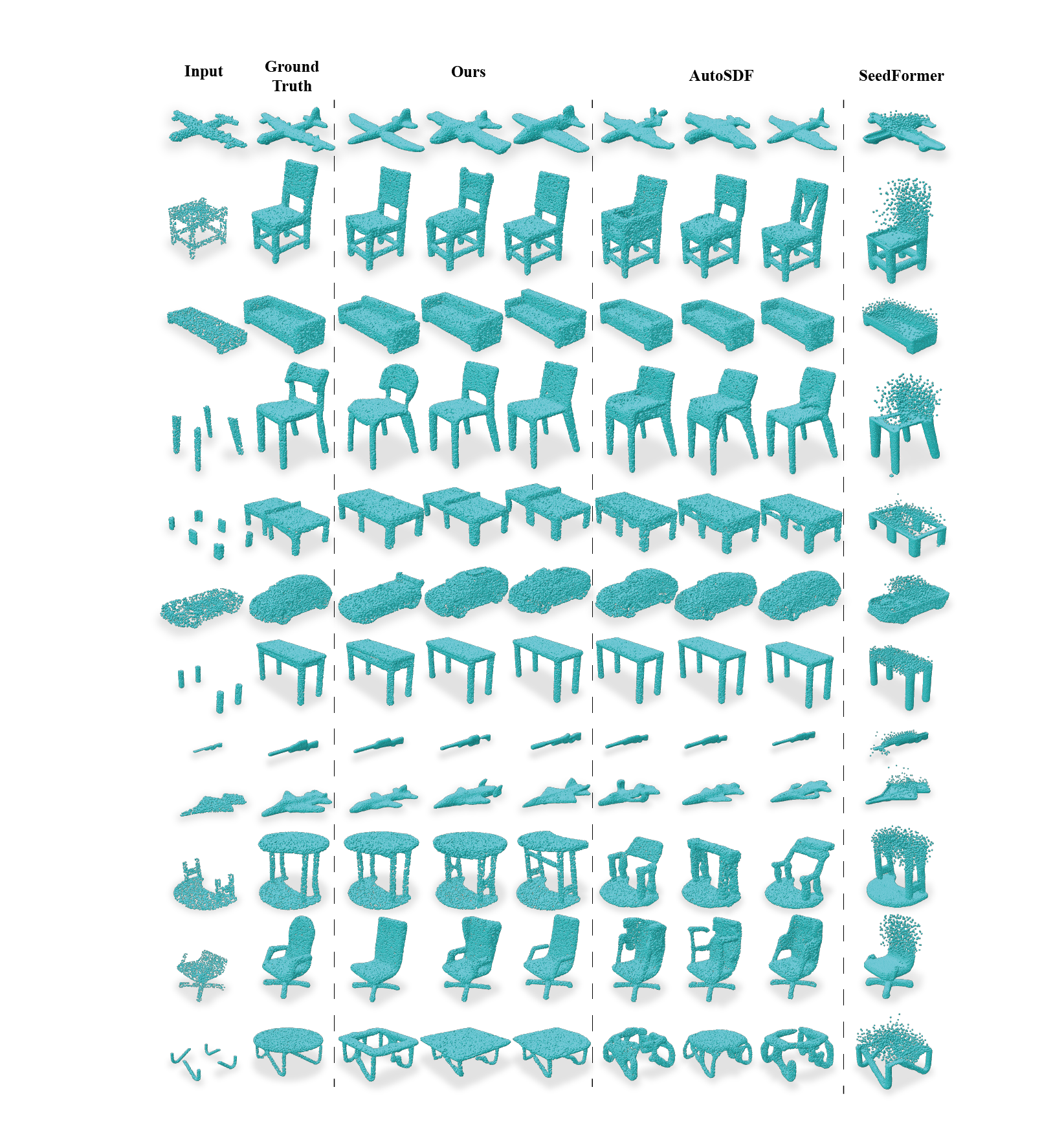}
    \vspace{-10mm}
   \caption{Visual comparison about shape completion given \textit{half} shapes.}
   \label{Fig: supp_half}
\end{figure*}

\begin{figure*}[]
  \centering
   \includegraphics[width=0.9\linewidth]{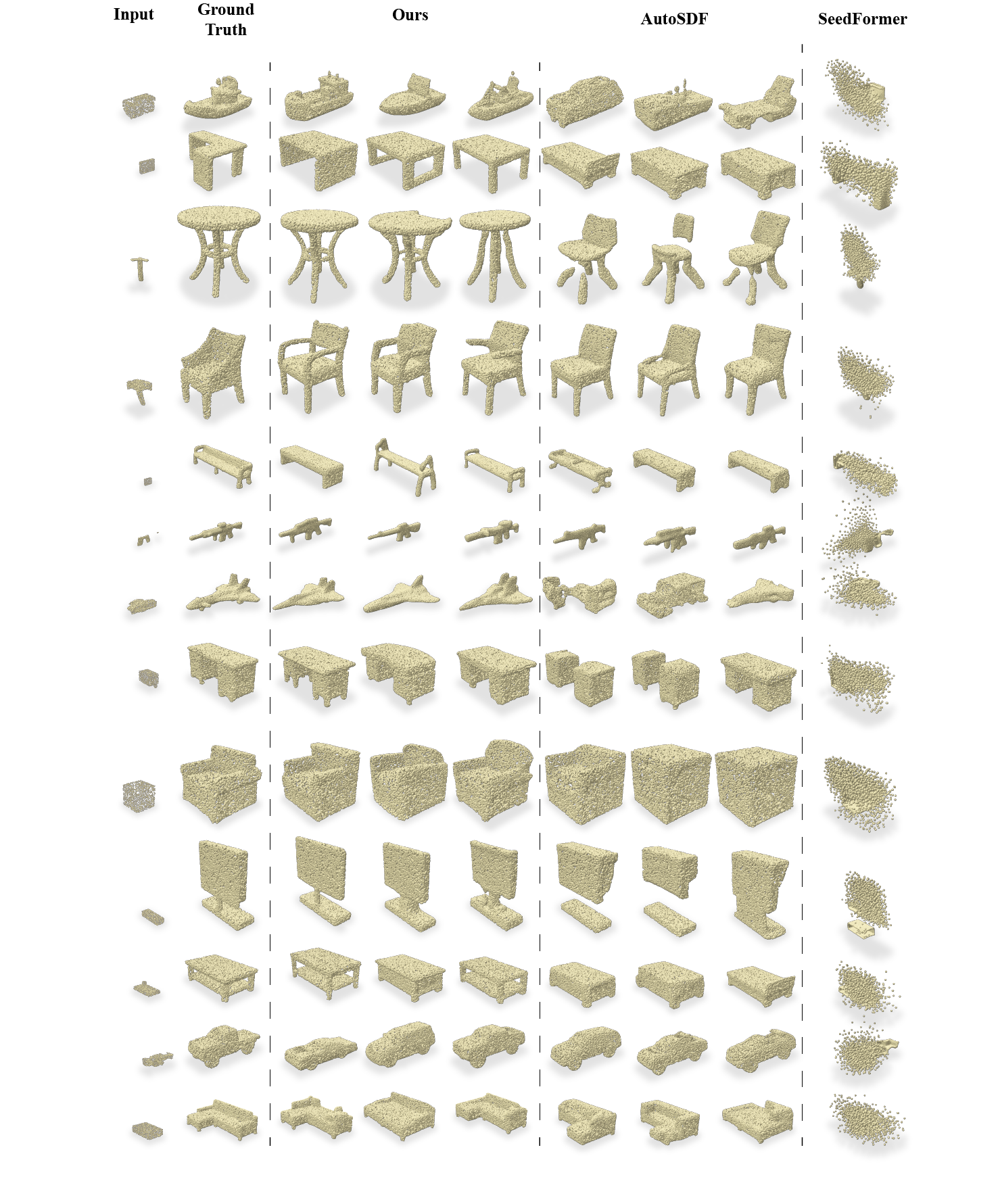}
    \vspace{-2mm}
   \caption{Visual comparison about shape completion given \textit{octant} shapes.}
   \label{Fig: supp_octant}
\end{figure*}

\begin{figure*}[]
  \centering
   \includegraphics[width=0.9\linewidth]{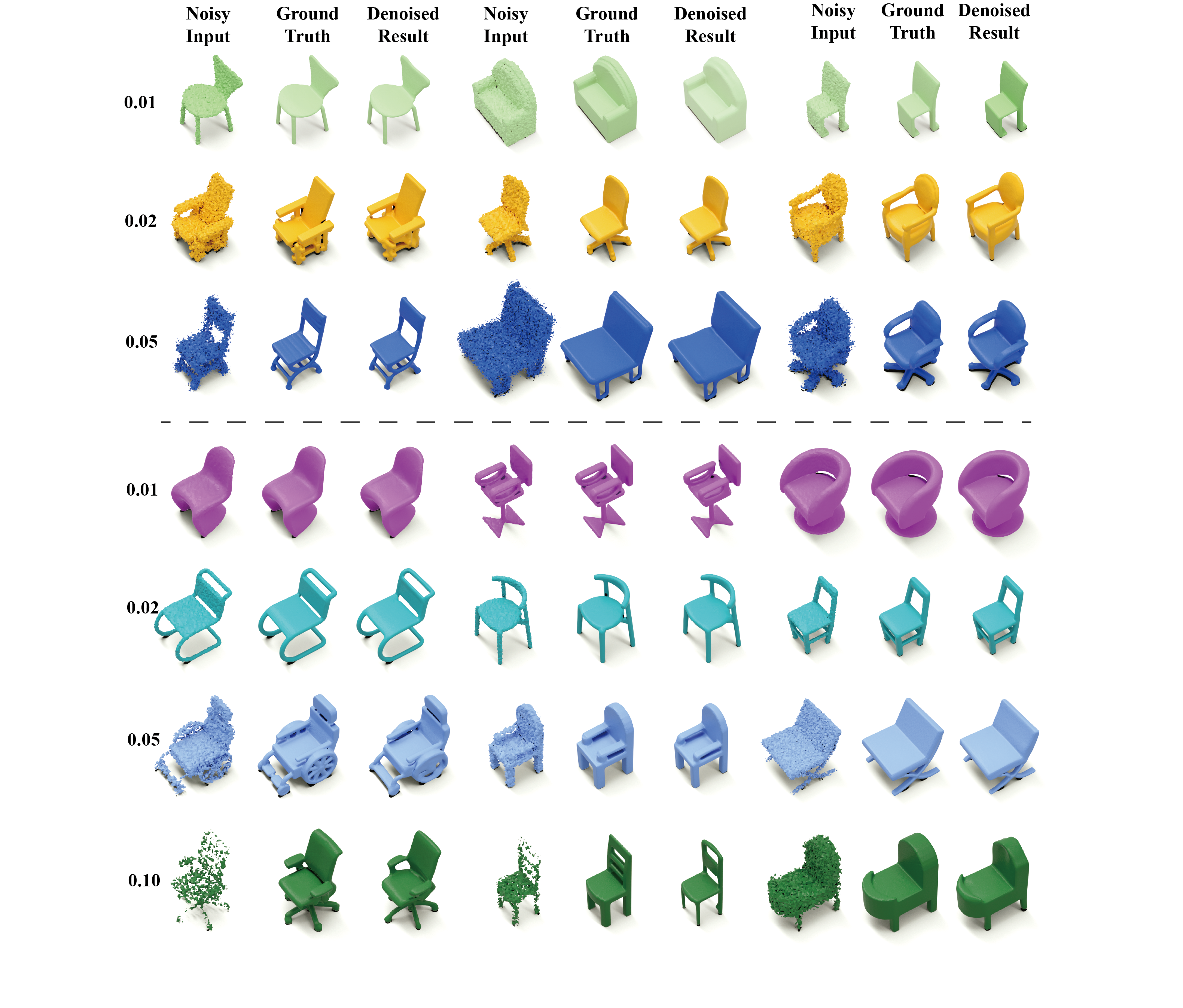}
    \vspace{-7mm}
   \caption{Visual results about denoising with various noise types and levels. \textit{Upper}: Gaussian noise. \textit{Lower}: Uniform noise.}
   \label{Fig: supp_denoise}
\end{figure*}

\begin{figure*}[]
  \centering
   \includegraphics[width=0.8\linewidth]{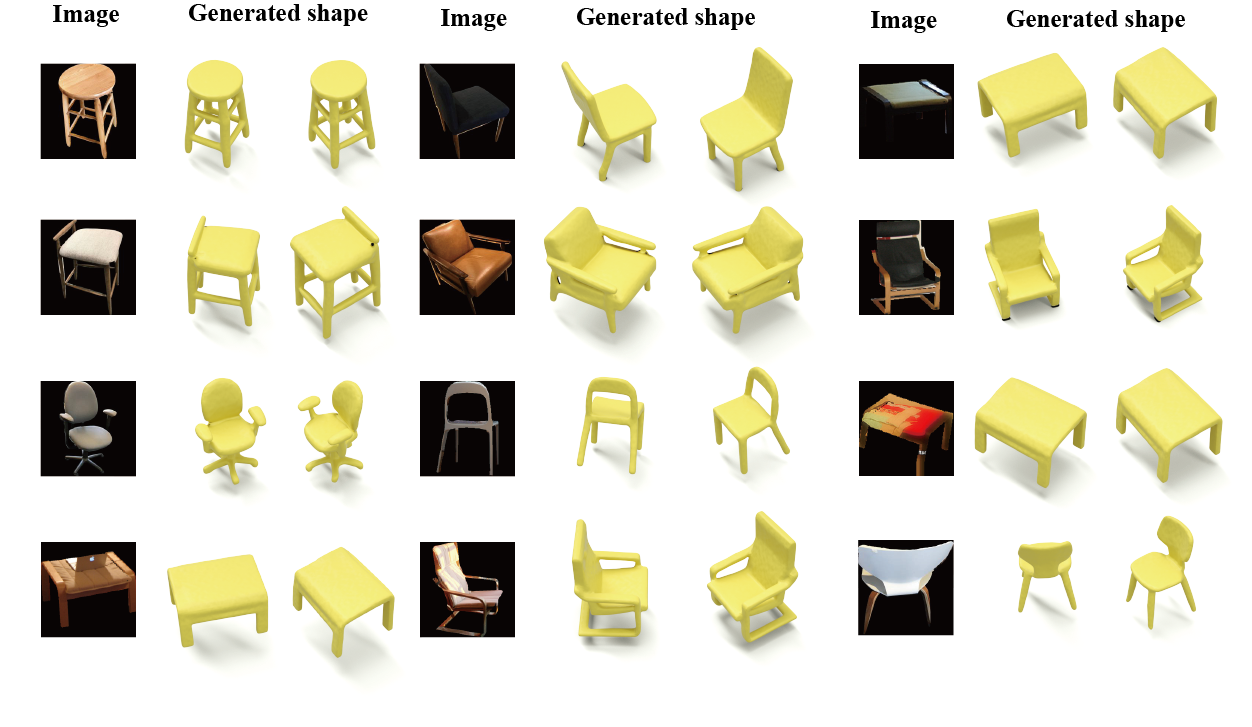}
    \vspace{-7mm}
   \caption{Visual results about single-view reconstruction.}
   \label{Fig: supp_img}
\end{figure*}

More visual results are presented in this section, including unconditional shape generation in~\cref{Fig: supp_uncond}, shape completion comparison in~\cref{Fig: supp_half} and~\cref{Fig: supp_octant}, denoising results in~\cref{Fig: supp_denoise}, single-view reconstruction in~\cref{Fig: supp_img}.

%%%%%%%%% REFERENCES
\clearpage
{\small
\bibliographystyle{ieee_fullname}
\bibliography{egbib}
}

%% file: cvpr2023-author_kit-v1_1-1/table/full_uncond.tex
\begin{table*}[]     
\centering          
\small              
\setlength{\tabcolsep}{6pt}
                    
\begin{tabular}{lcccccccccccc}
\toprule            
                    &                                                &\multicolumn{2}{c}{MMD$\downarrow$}&\multicolumn{2}{c}{COV$\uparrow$ (\%)}&\multicolumn{2}{c}{1-NNA$\downarrow$ (\%)} \\ \cmidrule(lr){3-4} \cmidrule(lr){5-6} \cmidrule(lr){7-8}
Category            &Model                                           &CD                  &EMD                 &CD                  &EMD                 &CD                  &EMD         \\ \midrule
                    &r-GAN~\cite{achlioptas2018learning}             &0.447              &2.309              &30.12               &14.32               &99.84               &96.79     \\        
Airplane            &l-GAN (CD)~\cite{achlioptas2018learning}        &0.340              &0.583              &38.52               &21.23               &87.30               &93.95          \\   
                    &l-GAN (EMD)~\cite{achlioptas2018learning}       &0.397              &0.417              &38.27               &38.52               &89.49               &76.91\\             
                    &PointFlow~\cite{yang2019pointflow}              &0.224              &0.390              &47.90               &46.41               &75.68               &70.74  \\           
                    &SoftFlow~\cite{kim2020softflow}                 &0.231              &0.375              &46.91               &47.90               &76.05               &65.80  \\           
                    &SetVAE~\cite{kim2021setvae}                     &0.200              &0.367              &43.70               &48.40               &76.54               &67.65 \\            
                    &DPF-Net~\cite{klokov2020discrete}               &0.264              &0.409              &46.17               &48.89               &75.18               &65.55 \\            
                    &DPM~\cite{luo2021dpm}                     &0.213                &0.572                  &48.64      &33.83               &76.42               &86.91  \\           
                    &PVD~\cite{zhou2021pvd}                        &0.224              &0.370              &\textbf{48.88}               &\textbf{52.09}      &73.82               &64.81 \\   \cmidrule{2-8}         
                    &3DQD \textit{(ours)}                                           &0.551             &0.426            &40.50              &47.17               &\textbf{56.29}      &\textbf{54.78}        \\
                    
\midrule              
                    &r-GAN~\cite{achlioptas2018learning}             &5.151               &8.312               &24.27               &15.13               &83.69               &99.70          \\   
Chair               &l-GAN (CD)~\cite{achlioptas2018learning}        &2.589               &2.007               &41.99               &29.31               &68.58               &83.84      \\       
                    &l-GAN (EMD)~\cite{achlioptas2018learning}       &2.811               &1.619               &38.07               &44.86               &71.90               &64.65\\             
                    &PointFlow~\cite{yang2019pointflow}              &2.409               &1.595               &42.90               &50.00               &62.84               &60.57     \\        
                    &SoftFlow~\cite{kim2020softflow}                 &2.528               &1.682               &41.39               &47.43               &59.21               &60.05\\             
                    &SetVAE~\cite{kim2021setvae}                     &2.545               &1.585               &46.83               &44.26               &55.84               &60.57\\             
                    &DPF-Net~\cite{klokov2020discrete}               &2.536               &1.632               &44.71               &48.79               &62.00               &58.53 \\            
                    &DPM~\cite{luo2021dpm}                     &2.399                &2.066                &44.86               &35.50                &60.05               &74.77 \\            
                    &PVD~\cite{zhou2021pvd}                        &2.622               &1.556               &\textbf{49.84}      &\textbf{50.60}              &56.26               &53.32         \\    \cmidrule{2-8}
                    &3DQD \textit{(ours)}                            &2.057               &1.128               &46.76               &48.13      &\textbf{55.61}      &\textbf{52.94}  \\  
                    
\midrule              
                    &r-GAN~\cite{achlioptas2018learning}             &1.446               &2.133               &19.03               &6.539               &94.46               &99.01    \\         
Car                 &l-GAN (CD)~\cite{achlioptas2018learning}        &1.532               &1.226               &38.92               &23.58               &66.49               &88.78    \\         
                    &l-GAN (EMD)~\cite{achlioptas2018learning}       &1.408               &0.899              &37.78               &45.17               &71.16               &66.19 \\            
                    &PointFlow~\cite{yang2019pointflow}              &0.901      &0.807              &46.88               &50.00               &58.10               &56.25   \\          
                    &SoftFlow~\cite{kim2020softflow}                 &1.187               &0.859              &42.90               &44.60               &64.77               &60.09    \\         
                    &SetVAE~\cite{kim2021setvae}                     &0.882               &0.733              &49.15               &46.59               &59.94               &59.94   \\          
                    &DPF-Net~\cite{klokov2020discrete}               &1.129               &0.853              &45.74               &49.43               &62.35               &54.48    \\         
                    &DPM~\cite{luo2021dpm}                     
                                                                     & 0.902  &1.140 &44.03  &34.94               &68.89               &79.97  \\           
                    &PVD~\cite{zhou2021pvd}                        &1.077               &0.794              &41.19               &50.56               &\textbf{54.55}             &53.83  \\           \cmidrule{2-8}
                    &3DQD \textit{(ours)}                                          &0.677               &0.443     &\textbf{49.28}      &\textbf{55.43}     &55.75      &\textbf{52.80}  \\  
                    
\midrule              
                    
\end{tabular}       
\caption{Generation performance metrics on Airplane, Chair, Car. MMD-CD is multiplied with $1\times 10^{3}$, MMD-EMD is multiplied with $1\times 10^{2}$. }
\label{tab:full_uncond}    
\end{table*}

%% file: cvpr2023-author_kit-v1_1-1/table/denoise_tab.tex
\begin{table}[]     
\centering          
\small              
\setlength{\tabcolsep}{6pt}
                    
\begin{tabular}{llccc}
\toprule            

Noise type          &Level            &MMD$\downarrow$      &AMD$\downarrow$     &TMD$\uparrow$     \\ \midrule
                    &0.01             &0.710                &0.897               &0.331         \\        
\textbf{Gaussian}   &0.02             &0.737                &0.918               &0.343         \\
                    &0.05             &1.006                &1.369               &0.543        \\          
                    &0.1              &3.680                &5.043               &1.724         \\
\midrule              
                    &0.01             &0.720                &0.906               &0.337         \\        
\textbf{Uniform}    &0.02             &0.756                &0.964               &0.355         \\
                    &0.05             &1.086                &1.448               &0.449        \\          
                    &0.1              &2.385                &3.167               &0.772         \\
                    
\midrule              
                    
\end{tabular}       
\caption{Shape denoising performance with different types and levels of noise. MMD and AMD is multiplied with $1\times 10^{3}$, TMD is multiplied with $1\times 10^{2}$. }
\label{tab:denoise}    
\end{table}         